# MVRS: The Multimodal Virtual Reality Stimuli-based Emotion Recognition Dataset

Seyed Muhammad Hossein Mousavi
C.E.O. and Researcher at Cyrus Intelligence Research Ltd
Tehran, Iran
ORCID ID: 0000-0001-6906-2152
s.m.hossein.mousavi@cyrusai.ir and mosavi.a.i.buali@gmail.com

Atiye Ilanloo
Independent Researcher
Rasht, Gilan, Iran
ORCID ID: 0009-0004-7087-3090
elanlooatiye@gmail.com

*Abstract*

**A**utomatic emotion recognition gained significant importance in the recent decade, especially with the development of artificial intelligence, which has affected our daily lives. Using personalized emotion recognition in healthcare, education, retail, and automotive has high importance these days, which requires proper data in different modalities. On the other hand, data scarcity in some emotion recognition modalities, such as body motion and physiological signals, is vivid, especially when it comes to multimodality. Furthermore, the way a participant is provoked to show emotion is crucial, as it should resemble real-life emotional expression. To do so, one of the most effective methods is to employ Virtual Reality (VR) videos and games. This paper introduces a novel Multimodal Virtual Reality Stimuli-based emotion recognition dataset, or MVRS, which could address the mentioned data scarcity issue. Our dataset contains 13 subjects or participant stimuli using VR videos for relaxation, fear, stress, sadness, and joy emotions. The dataset covers an age range of 12 to 60 in both genders. This dataset is recorded in a small lab in which all participants followed the same data collection protocols and filled out both questionnaires and consent forms. The dataset includes eye tracking, body motion, ElectroMyoGraphy (EMG), and Galvanic Skin Response (GSR) data in various formats. The eye-tracking data is recorded using a Full High-Definition (FHD) webcam placed manually into the VR Head-Mounted Display (HMD). The body motion data is recorded using Microsoft Kinect version 2. Finally, EMG and GSR data are recorded by an Arduino UNO board. All data is recorded simultaneously, with synchronized timestamps, to support clean data for multimodal processing. For each modality, related features are extracted and fused by multimodal fusion techniques (early and late stages) and evaluated using different classifiers and metrics to check the validity and separability of the data.

**K**eywords: Emotion Recognition, Virtual Reality, Multimodality, Eye Tracking, Arduino, ElectroMyoGraphy (EMG), Galvanic Skin Response (GSR)

- # Introduction
  - ### Basic Definitions and Challenges

Emotion recognition is the computational process by which systems infer human emotional states such as joy, fear, sadness, and stress by analyzing observable behavioral cues (e.g., body posture and motion [24], facial gestures [26]) and involuntary physiological signals [1, 2, 25]. This technology has seen significant progress and is being used in diverse domains including healthcare (e.g., stress monitoring, mental health assessment), education (e.g., adaptive tutoring), automotive (e.g., detecting driver fatigue), and immersive Human–Computer Interaction (HCI) [23] (e.g., emotion-aware Virtual Reality (VR) environments) [2, 3]. Behavioral data captured via Red Green Blue-Depth (RGB-D) [28] sensors like Microsoft Kinect[1] or wearable Inertial Measurement Units (IMUs) provide powerful, real-time indicators of affective state from body movements [4, 24], while physiological modalities offer reliable internal measures: ElectroMyoGraphy (EMG) and ElectroCardioGram (ECG) signal muscular and cardiac activity; Galvanic Skin Response (GSR) reflects autonomic arousal; respiration depth and rate map to stress and relaxation levels [2, 5]. Multimodal fusion approaches that integrate body motion and physiological inputs, using early or late fusion, have consistently been shown to outperform unimodal systems by combining contextual expressivity with involuntary authenticity [2, 5, 6]. Despite these advances, several challenges persist: first, the scarcity of large-scale datasets with synchronized body and physiological signals, especially for spontaneous emotional states, limits model generalizability [4, 3]. Second, physiological signals exhibit significant inter-subject variability and are prone to noise and motion artifacts, complicating preprocessing and classifier training [5, 6]. Third, aligning and fusing heterogeneous data streams (e.g., timestamp synchronization of EMG, video, and GSR) demands robust architectures and precise calibration [6, 2]. Lastly, eliciting ecologically valid emotions in controlled settings remains difficult: traditional stimuli (e.g., images or fixed videos) often fall short in evocation, while VR has emerged as a promising tool but still requires more validation through controlled dataset collection [3, 4]. Collectively, these challenges underscore the urgent need for new, multimodal datasets, such as our proposed dataset, that record synchronized body motion and physiological data under immersive VR conditions to support the development of more accurate, robust emotion recognition systems. Due to the nature of our dataset, we name it the Multimodal Virtual Reality Stimuli-based emotion recognition dataset (MVRS).

---

[1] https://azure.microsoft.com/en-us/products/kinect-dk

❖ **Importance and Motivation**

Emotion recognition holds crucial importance in modern intelligent systems, offering manifold motivations for its development. First, understanding emotional states through body motion and physiological signals is vital for healthcare applications, including real-time stress detection, mental health monitoring, and adaptive therapy, especially when evaluated in immersive environments like VR that induce more authentic responses [7, 8]. Second, in human-computer interaction, emotion-aware systems enable more empathetic and responsive interfaces, adapting tutoring systems, gaming platforms, and virtual assistants in real-time to users' feelings, thus enhancing engagement and usability [8, 9]. Third, the advent of multimodal fusion methods that use ECG, EMG, GSR, respiration, gaze, and motion data has demonstrated significantly higher accuracy and robustness over unimodal approaches, reinforcing the motivation to integrate diverse signals [7, 10, 11]. Fourth, VR-based emotion elicitation frameworks notably improve the ecological validity of datasets; real immersion elicits genuine physiological and motion responses essential for generalizable model training [9, 12]. Finally, robust emotion recognition systems have the potential to revolutionize domains such as automotive safety (e.g., monitoring driver fatigue), mental well-being diagnostics, and personalized learning by catering systems to individual affective states [8, 9]. Consequently, there is a strong impetus to develop well-synchronized, multimodal emotion datasets, especially those incorporating immersive VR stimuli, to drive innovation in emotion-aware technologies.

❖ **Virtual Reality (VR)**

Virtual Reality (VR) is an immersive technology that creates a simulated, interactive environment enabling users to experience and manipulate 3-Dimensional (3D) worlds that mimic or differ from real life [13]. By engaging multiple senses, VR has found important applications in fields such as healthcare, for pain management, therapy, and rehabilitation, education for experiential learning [23], entertainment and gaming, and training simulations for high-risk professions [14, 15, 27]. In emotion recognition research, VR provides a uniquely controlled yet ecologically valid environment that can elicit authentic emotional responses, making it a powerful tool for generating naturalistic affective data [16, 17]. Despite its advantages, VR also faces challenges: hardware limitations such as latency and field of view can reduce immersion quality; user discomfort and cybersickness remain significant issues; and high costs and setup complexity limit accessibility for widespread use [18]. Additionally, creating emotionally compelling and ethically sound VR stimuli requires interdisciplinary collaboration and careful experimental design to avoid stress or harm to participants [16]. Nonetheless, VR's capacity to simulate real-world scenarios in a repeatable and measurable manner continues to drive its growing adoption across research and industry domains.

❖ **Multimodality**

Multimodality in emotion recognition involves integrating data from multiple sources or sensory channels, such as facial expressions, body motion, speech, physiological signals, gaze, and more, to capture the complex, multifaceted nature of human affect [1]. Multimodal datasets that combine these heterogeneous signals have become essential for developing robust affective computing systems, as they provide complementary information that overcomes the limitations inherent in unimodal data [19]. For example, physiological signals offer internal, involuntary markers of emotional states, while behavioral cues like facial expressions and gestures provide contextual and expressive details, which together enhance classification accuracy and reliability [20, 2]. Studies consistently show that multimodal fusion methods[2] outperform unimodal approaches by effectively using the strengths of each modality, resulting in improved generalization and robustness to noise and missing data [19, 21]. This superiority is particularly important in real-world applications such as healthcare monitoring, adaptive learning, HCI, and immersive virtual environments [23], where emotional signals may vary widely in quality and availability [22]. The creation of synchronized multimodal datasets with naturalistic emotional elicitation protocols thus remains a critical step to advance emotion recognition research and to facilitate the development of practical, scalable emotion-aware systems [1, 20]. Figure 1 depicts a conceptual view of multimodal fusion.

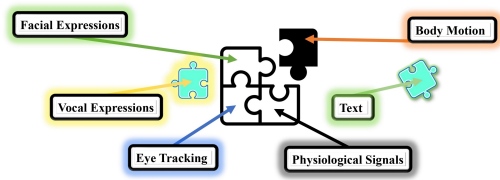

Figure 1. The Conceptual View of Multimodal Fusion

---

[2] https://github.com/SeyedMuhammadHosseinMousavi/Multi-Modal-Fusion

- **Eye Tracking Modality**

Eye tracking is a non-invasive and informative modality that captures gaze direction, fixation duration, saccades, and pupil dilation to infer a user's cognitive and emotional states[3]. Because eye movements are often subconscious, they serve as reliable indicators of arousal, attention, and affective engagement, making them valuable for emotion recognition systems. For instance, variations in fixation patterns and pupil diameter have been linked to emotional intensity and valence during the viewing of affective stimuli [29]. Moreover, when combined with immersive environments such as VR, eye tracking enables dynamic, context-aware affect analysis by mapping gaze behavior directly onto interactive content [3]. This allows for the development of adaptive systems that respond to emotional cues in real time, improving applications in education, therapy, and user experience research. Figure 2 illustrates a frame sample of eye tracking with some processing on it, belonging to one of our participants. The eye is a bit closed and the pupil is a bit opener than usual, which indicates mild sadness.

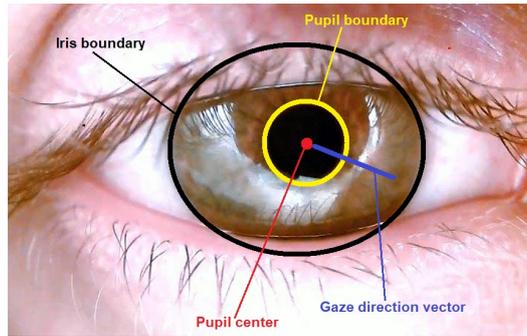

Figure 2. An eye-tracking frame sample of our MVRS dataset and some basic processing as annotations (sadness)

- **Body Motion Modality**

Body motion is a rich and expressive modality for emotion recognition, as human emotions are often communicated through nonverbal cues such as posture, gestures, and movement dynamics. Body motion is a representation of joints' position and rotation in the 3D space over frames[4]. Unlike facial expressions, which may be consciously controlled or masked, body movements tend to reflect emotional states more naturally and holistically, particularly during full-body interactions or spontaneous behavior. Studies have shown that features like joint angles, movement speed, symmetry, and expansiveness correlate strongly with specific emotional categories such as sadness, anger, or joy [30, 31]. In real-time applications, motion capture systems like Microsoft Kinect or wearable IMUs enable the tracking of skeletal joint data, which can be processed for temporal and spatial patterns indicative of affective state [4]. As such, body motion offers a powerful and intuitive channel for emotion recognition, especially in scenarios where facial visibility is limited or where whole-body interaction is natural, such as in VR or embodied agent environments. Figure 3 shows a depth sample image of our dataset, having a skeleton structure for body motion processing. Body joints are steady, not expanded nor bended, which indicates neutrality or relaxation, which in this case is relaxation.

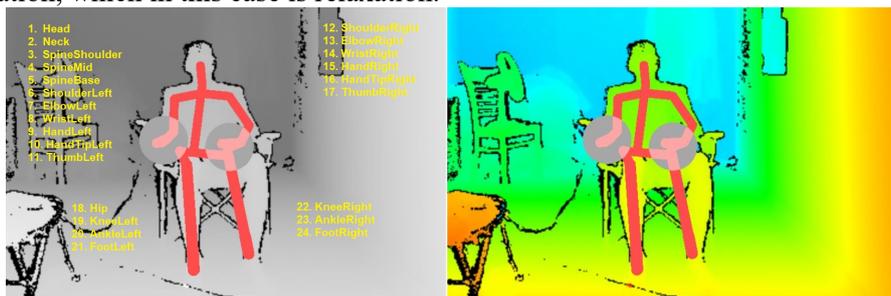

Figure 3. A depth sample view of our MVRS dataset in depth gray and colored formats, having some basic body joint annotations (relaxation)

- **Physiological Signals Modality**

Physiological signals offer a direct window into the autonomic and somatic nervous systems, providing robust indicators of emotional arousal and intensity[5]. Among these, EMG measures electrical activity produced by skeletal muscles and is particularly effective for detecting subtle facial expressions or tension-related muscle activations associated with emotional states such as stress, fear, or joy. On the other hand, GSR, also known as ElectroDermal Activity (EDA), tracks skin conductance changes driven by sweat gland activity, offering a reliable marker of sympathetic nervous system arousal. GSR is especially useful in distinguishing between high and low arousal states, making it a cornerstone in affective computing applications [32]. When used together, EMG and GSR can capture both voluntary and involuntary emotional responses, and their integration enhances the accuracy of emotion recognition systems in real-time contexts such as HCI, biofeedback, and VR-based studies [33]. Their low cost, ease of use, and high temporal resolution make them attractive modalities for multimodal emotion recognition research. Figure 4 depicts parts of the GSR and EMG time-series data with the same timestamps belonging to one of the experiments under stress stimuli. The figure shows a rise in the GSR values and fluctuation in the EMG ones, which indicates uncomfortable states such as fear or stress, which in this case is stress.

---

[3] https://dac.digital/what-is-eye-tracking-technology-and-how-does-it-transform-different-industries/

[4] https://github.com/SeyedMuhammadHosseinMousavi/Graph-Based-Parallel-Multi-Objective-Optimization-of-Skeletal-Body-Motion-Data

[5] https://www.eecs.qmul.ac.uk/mmv/datasets/deap/

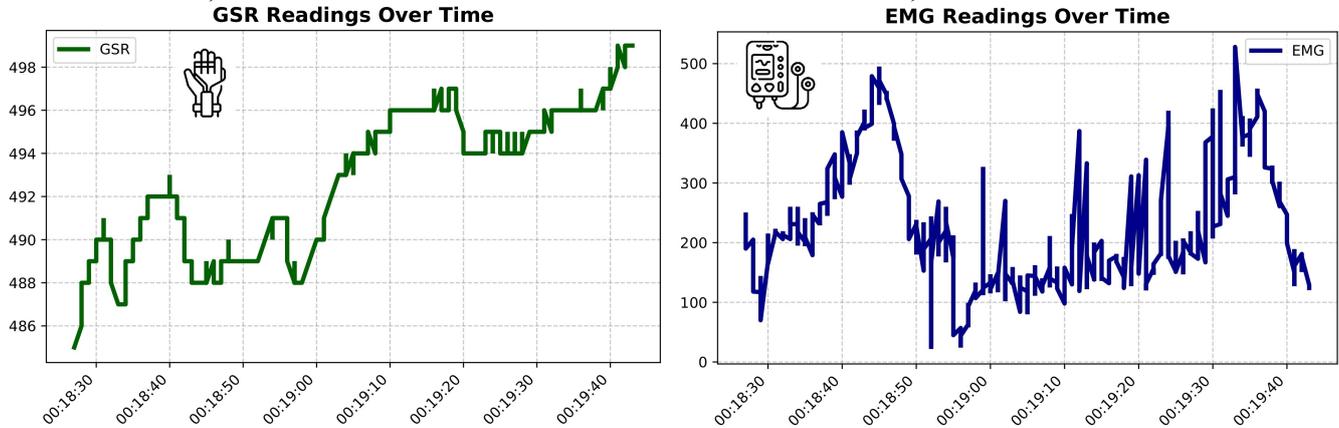

Figure 4. Portion of the time-series data belonging to one of the participants in our MVRS dataset during the stress stimuli video, with the same timestamps

❖ **Contribution**

In this study, we propose a novel Multimodal Virtual Reality Stimuli-based emotion recognition dataset (MVRS) to address the growing need for diverse and realistic affective data, particularly in underrepresented modalities such as body motion and physiological signals. Unlike many existing datasets that rely on traditional stimuli or lack synchronized multimodal acquisition, our dataset is designed around immersive VR environments, which elicit more naturalistic and lifelike emotional expressions. The MVRS dataset includes recordings from 13 participants across five emotional states: relaxation, fear, stress, sadness, and joy, using carefully curated VR videos. It spans a diverse demographic (ages 12 to 60, across both binary genders) and captures simultaneous signals across multiple modalities: body motion (via Microsoft Kinect v2), EMG and GSR (via Arduino UNO), and eye-tracking (via a manually mounted Full HD webcam inside the VR headset). All signals are synchronized with precise timestamping to ensure temporal alignment across modalities, making the dataset highly suitable for multimodal fusion tasks. In addition, each participant completed standardized emotion questionnaires and signed ethical consent forms to validate ground truth labels. We further extract relevant features from each modality, including statistical, temporal, and frequency-domain descriptors, and apply early and late multimodal fusion techniques to assess emotion separability. The quality and discriminative power of the dataset are evaluated using several classifiers (Random Forest, Decision Tree, XGBoost) and metrics, showcasing its effectiveness for both unimodal and multimodal emotion recognition pipelines. Our contributions lie in (1) designing a synchronized, naturalistic VR-based emotion elicitation protocol, (2) collecting a rich and diverse multimodal dataset, and (3) establishing a complete baseline framework for emotion classification through multimodal fusion and traditional machine learning approaches. This work provides a valuable resource for advancing affective computing, particularly in healthcare, education, and adaptive HCI applications. A point is that during the paper, terms like stimulus, elicitation, and induce are used for the same application.

❖ **Research Questions**

We aim to answer the following Research Questions (RQs) by producing this dataset. How effective are VR environments in eliciting naturalistic emotional responses compared to traditional 2D stimuli in emotion recognition tasks? To what extent can synchronized multimodal data, including GSR, EMG, body motion, and eye-tracking, enhance the accuracy and reliability of automatic emotion recognition models? What is the comparative impact of early fusion versus late fusion strategies on the performance of emotion classification in multimodal settings? How do different machine learning classifiers (e.g., Random Forest, Decision Tree, XGBoost) perform across single-modality versus fused-modality datasets in distinguishing among relaxation, fear, stress, sadness, and joy? What are the most discriminative features across each modality (GSR, EMG, motion, eye tracking) that contribute most significantly to emotion classification in a VR-based setup?

❖ **Paper Structure**

The structure of this paper is organized to guide the reader from foundational concepts to the contributions and evaluation of our proposed system. The introduction provides an overview of definitions, the importance of emotion recognition in modern applications such as healthcare and education, the challenges of multimodal data scarcity, and the motivation behind developing a VR-based dataset. Following that, the literature section reviews existing unimodal and multimodal emotion recognition datasets to underline the research gap our work aims to address. The construction of the proposed MVRS dataset section details the development process of the dataset, including participant information, the VR stimuli used, hardware setups (VR HMD, Kinect, Arduino), types of data collected, standardized data collection protocols, and practical challenges encountered during acquisition. The Evaluation of the MVRS dataset section explains the techniques used for data preprocessing, feature extraction, and classification, along with the metrics used to validate the dataset's effectiveness and emotional separability. The conflict of interest statement emphasizes that the research and dataset collection were solely conducted and funded by the authors under the Cyrus Intelligent Research initiative. Finally, the conclusion summarizes the key findings, reflects on the impact of the MVRS dataset, and proposes future research directions in VR-based multimodal emotion recognition.

- **Literature**
  - ❖ **Unimodal and Multimodal Emotion Recognition Datasets**
    - o **Image-based Emotion Recognition Datasets**

The FER-2013 dataset[6] [34] was introduced during the ICML 2013 challenges in representation learning. It contains 35,887 grayscale facial images of size 48×48 pixels labeled with seven emotion classes: anger, disgust, fear, happiness, sadness, surprise, and neutral. This dataset is widely used as a benchmark for emotion classification tasks, especially in deep learning models, due to its large scale and challenging real-world variability. AffectNet[7] [35] is one of the largest facial expression datasets in the wild, consisting of over 1 million facial images collected from the internet. It is annotated with both categorical emotions (8 classes: happy, sad, angry, fear, disgust, surprise, contempt, and neutral) and dimensional emotion labels (valence and arousal). Its diversity and inclusion of spontaneous expressions make it suitable for robust affective computing research. The Oulu-CASIA[8] [36] dataset provides both near-infrared and visible light facial images recorded from 80 subjects expressing six emotions. The recordings are captured under different lighting conditions, making the dataset valuable for evaluating expression recognition models in varying environments. It includes sequences of dynamic facial expressions that transition from neutral to peak expression. The BU-3DFE[9] [37] is a 3D facial expression dataset that includes textured 3D scans from 100 subjects (56% female, 44% male) across six prototypical emotions at four intensity levels. It provides both geometry and texture information, making it ideal for depth-based and 3D emotion recognition research. The high-fidelity expression annotations help in studying subtle emotional cues in 3D space. The Iranian Kinect Face Database (IKFDB) [38] is an RGB-D facial expression dataset collected using Kinect v2. It includes 40 participants aged 7 to 70, with over 100,000 synchronized color and depth frames covering five basic emotions. The dataset emphasizes diverse ethnic representation and supports both macro- and micro-expression analysis. It has been evaluated using HOG features and classifiers like SVM and CNN, showing competitive accuracy with existing benchmarks. This dataset is available upon request.

   - o **Physiological Signal-based Emotion Recognition Datasets**

The DEAP dataset[10] [2] includes physiological and ElectroEncephaloGram (EEG) recordings from 32 participants who watched 40 music video clips designed to elicit emotions. Although it contains multiple modalities, many studies focus only on EEG or a single peripheral signal like GSR, making it applicable in unimodal settings. It includes self-reported valence, arousal, dominance, and liking scores. The DREAMER dataset[11] [39] contains EEG and ECG recordings from 23 participants exposed to audiovisual emotional stimuli. Researchers often use only the EEG or ECG from this dataset to study emotion in a unimodal manner. The dataset is labeled with self-assessed valence, arousal, and dominance. The MAHNOB-HCI[12] [40] is a multimodal dataset captured from 30 participants during emotion-eliciting video sessions. For unimodal studies, GSR or ECG signals are commonly extracted and used independently. Participants also provided self-reports of emotional responses after each video. The BioVid EmoDB dataset[13] [41] was collected to investigate physiological responses to emotional and pain stimuli using GSR, ECG, and EMG. Although multiple signals were collected, many works isolate one modality, such as GSR or EMG, to evaluate emotional states. Stimuli were presented in a structured lab setting with graded levels of emotional intensity. The WESAD dataset[14] [42] involves wearable sensor data from 15 subjects during baseline, stress, and amusement states. It includes GSR, ECG, temperature, and accelerometer data from chest and wrist devices. Often, only GSR or ECG is used in unimodal emotion recognition studies. The AVDOS-VR dataset[15] [43] is a publicly available VR-based video emotion elicitation database that includes continuous self-ratings of valence and arousal alongside physiological signals collected using an emteqPRO mask integrated with a VR HMD, encompassing PPG, facial EMG (7 channels), and IMU data. It comprises recordings from 37 participants who watched 30 standardized 30-second affective videos (10 positive, neutral, negative), with built-in 2-minute relaxation intervals. Data collection was done in both home and lab environments remotely using guided VR setups, demonstrating ecological validity. This dataset supports research into immersive affective computing and remote emotion monitoring, bridging a gap in VR-based physiological emotion datasets.

---

[6] https://www.kaggle.com/datasets/msambare/fer2013
[7] https://huggingface.co/datasets/chitradrishti/AffectNet
[8] https://www.oulu.fi/en/university/faculties-and-units/faculty-information-technology-and-electrical-engineering/center-for-machine-vision-and-signal-analysis
[9] https://cove.thecvf.com/datasets/529
[10] https://www.eecs.qmul.ac.uk/mmv/datasets/deap/
[11] https://zenodo.org/records/546113
[12] https://torcheeg.readthedocs.io/en/v1.1.0/generated/torcheeg.datasets.MAHNOBDataset.html
[13] https://www.nit.ovgu.de/nit/en/BioVid-p-1358.html

  o **Eye Tracking-based Emotion Recognition Datasets**

The VREED[16] [3] is a multimodal dataset where emotions were triggered using immersive 360° video-based virtual environments delivered via a VR headset. Behavioral (eye tracking) and physiological signals, ECG and GSR, were captured, together with self-reported responses, from 34 participants experiencing 12 VR scenarios. Statistical analysis confirmed the validity of the selected 360 in eliciting the desired emotions. Preliminary machine learning analysis demonstrated state-of-the-art performance reported in affective computing literature using non-immersive modalities. The eSEE-d dataset[17] [44] comprises eye-tracking data from 48 participants who viewed 10 emotion-evoking videos, each followed by a neutral video. Participants rated five emotions (tenderness, anger, disgust, sadness, neutral) on a scale from 0 to 10, later translated into emotional arousal and valence levels. The dataset includes 28 eye and gaze features such as fixation duration, saccade amplitude, blink frequency, and pupil diameter, enabling the classification of various arousal and valence levels based solely on eye and gaze features. Machine learning approaches demonstrated promising results, with a Deep Multilayer Perceptron (DMLP) network achieving 92% accuracy in distinguishing positive valence from non-positive and 81% in distinguishing low arousal from medium arousal. The EyeT4Empathy dataset[18] [45] focuses on assessing empathy through eye-tracking data. Participants were asked to use eye movements on a letterboard to write sentences, while a control group engaged in foraging for visual information. The dataset includes eye-tracking data collected during these tasks, along with pre- and post-intervention questionnaires assessing empathy levels. The data collected from the test group (gaze typing) and the control group (foraging) are stored separately, facilitating the study of empathy through eye behaviors

  o **Body Motion-based Emotion Recognition Datasets**

The MPI dataset[19] [56] includes over 1,400 motion capture recordings of eight actors performing emotional body expressions within narrative scenarios. It covers ten emotions, such as joy, sadness, anger, and pride, captured at 120 fps using a full-body motion capture system. Both intended (actor) and perceived (observer-rated) emotion labels are provided. It is specifically designed for research in body-language-based emotion recognition. The CMU-MMAC database[20] [57] contains multimodal recordings of subjects performing everyday cooking activities in a kitchen environment. Body motion data was captured using a Vicon motion capture system with 41 markers, alongside other modalities such as video, audio, and inertial data. While not solely emotion-centric, the dataset is valuable for recognizing naturalistic behaviors and can be used in affective computing contexts where emotional states are inferred from activity and body movement. It provides a realistic, complex setting with natural full-body movements and interactions with objects. The 100 styles dataset[21] [58] was developed for research on stylistic modeling of human locomotion. It contains motion capture recordings of human walking and running with variations in emotional style, such as happy, sad, or angry gaits. The focus is on full-body movement and stylistic differences across motion phases, captured using a marker-based motion capture system. Although the dataset was primarily used to develop real-time transformation and control models, its emotionally stylized motion makes it relevant for emotion recognition through body dynamics. The KDAEE dataset[22] [59] contains full-body motion capture recordings of professional actors performing scripted emotional expressions such as joy, sadness, anger, fear, disgust, and surprise. The kinematic data include joint positions and movements, captured using an optical motion capture system. The focus is on dynamic expression through body posture and gesture, excluding facial expressions or voice, making it a purely body-motion unimodal emotion recognition dataset. The dataset is valuable for studying emotion communication through nonverbal movement. The Pacolab (PACO-Lab) dataset[23] [60] contains 4,080 motion-capture recordings from 30 non-professional actors (15 female, 15 male). Each actor performed four action types (walking, knocking, lifting, and throwing) across four affective styles: angry, happy, sad, and neutral. Data were captured using a high-speed optical motion-capture system, with joint positions formatted for use with animation and kinematic analysis tools. This library enables research into how identity, gender, and emotion are expressed and perceived through human biological motion. The AMASS dataset[24] [61] is a comprehensive dataset that consolidates 15 motion capture datasets into a unified format by converting raw marker-based motion data. It includes over 40 hours of motion recordings across more than 11,000 sequences from over 300 subjects, capturing a wide range of movements including daily actions, expressive behaviors, and complex physical dynamics. The dataset provides high-quality 3D body meshes, joint positions, and soft tissue deformations, making it suitable for tasks such as human motion analysis, animation, and emotion recognition based on body motion. AMASS is widely used for research in modeling natural human movements across diverse scenarios. The BandaiNamco dataset[25] [62] was developed to support the training and evaluation of AI models for motion editing and stylization. It includes a wide range of human actions captured via optical motion capture systems, focusing on both natural and exaggerated motions suitable for stylization tasks. The dataset provides high frame rate 3D motion data annotated with action categories, enabling realistic motion synthesis, retargeting, and emotion-driven motion generation. It is especially designed to facilitate practical applications such as game animation, virtual avatars, and AI-based motion tools.

---

[14] https://ubi29.informatik.uni-siegen.de/usi/data_wesad.html
[15] https://www.gnacek.com/affective-video-database-online-study
[16] https://www.kaggle.com/datasets/lumaatabbaa/vr-eyes-emotions-dataset-vreed
[17] https://zenodo.org/records/5775674
[18] https://gts.ai/dataset-download/eyet4empathy-eye-movement-and-empathy-dataset/
[19] https://www.mpi-inf.mpg.de/departments/computer-vision-and-machine-learning/software-and-datasets/mpii-human-pose-dataset
[20] https://kitchen.cs.cmu.edu/
[21] https://www.ianxmason.com/100style/

- o **Speech-based Emotion Recognition Datasets**

The Emo-DB[26] is a German-acted speech dataset [46] featuring 10 actors (5 male, 5 female) who each recorded 535 utterances across seven emotions: anger, boredom, disgust, fear, happiness, sadness, and neutral. It's widely used for baseline evaluations in emotion classification. The SAVEE dataset[27] [47] includes emotional speech from four male British actors delivering 480 utterances labeled across seven emotions (anger, disgust, fear, happiness, sadness, surprise, and neutral). Although it has video data, it's frequently used in audio-only experiments. The CREMA-D[28] [48] contains 7,442 audio clips from 91 actors (43 female, 48 male) performing 12 sentences in six emotions: anger, disgust, fear, happiness, neutral, and sadness. The database is widely used in audio-only emotion classification research. The RAVDESS dataset[29] [49] features 24 actors (12 male, 12 female) with 1,440 speech and 1,080 song recordings across eight emotions: calm, happy, sad, angry, fearful, surprise, disgust, and neutral. Frequently used in audio-only analyses. The IEMOCAP dataset[30] [50] is a rich database with dyadic scripted and improvised dialogues from 10 actors, featuring both audio and motion capture data. For unimodal emotion recognition, only the speech audio is often used, with utterances labeled across emotions like anger, happiness, sadness, frustration, and neutral.

- o **Text-based Emotion Recognition Datasets**

The GoEmotions dataset[31] [51] is a fine-grained emotion dataset released by Google, consisting of 58,000 English Reddit comments annotated with 27 emotion labels (e.g., amusement, anger, gratitude, fear) plus a neutral class. It offers multi-label annotations where each comment can express more than one emotion. The annotations were done by human raters to ensure consistency, and the dataset is often used for benchmarking large language models in nuanced emotion classification. The SemEval-2007 affective text dataset[32] [52] was introduced as part of the SemEval-2007 shared task. It contains 1,250 English news headlines, each annotated for six Ekman-style basic emotions (anger, disgust, fear, joy, sadness, surprise). Emotion intensity is also included, enabling both classification and regression tasks. Headlines are short but emotionally rich, making this set ideal for evaluating emotional nuance in brief texts. The EmoBank dataset[33] [53] contains 10,000+ sentences with dual emotion annotations: categorical (Ekman's six emotions) and dimensional using Valence, Arousal, Dominance (VAD) scores. It enables nuanced emotion modeling and is suitable for tasks involving emotion intensity, affective computing, and sentiment regression. Each sentence comes from real-world texts and was annotated with crowd-sourced evaluations. The Daily Dialog dataset[34] [54] is a large-scale, high-quality dialogue dataset with over 102,000 utterances in 13,000 multi-turn conversations, all written in English. Each utterance is labeled with one of seven emotion categories (anger, disgust, fear, happiness, sadness, surprise, or "others") and also has labels for act types (e.g., question, statement). It's designed for building conversational agents that can recognize emotions in everyday human interactions. The EmoryNLP dataset[35] [55, 63] consists of 12,600+ utterances from Friends TV series dialogues, the sitcom, annotated with seven emotion labels (anger, disgust, fear, joy, sadness, surprise, neutral). It emphasizes contextual emotion recognition, i.e., the emotional interpretation of an utterance depending on previous dialogue. Due to its natural, conversational tone and character-based variation, it's used for training dialogue systems and emotion-aware models.

---

[22]

[23] https://paco.psy.gla.ac.uk/?page_id=14973

[24] https://amass.is.tue.mpg.de/

[25] https://github.com/BandaiNamcoResearchInc/Bandai-Namco-Research-Motiondataset

[26] https://www.kaggle.com/datasets/piyushagni5/berlin-database-of-emotional-speech-emodb

[27] https://www.kaggle.com/datasets/ejlok1/surrey-audiovisual-expressed-emotion-savee

[28] https://www.kaggle.com/datasets/ejlok1/cremad

[29] https://www.kaggle.com/datasets/uwrfkaggler/ravdess-emotional-speech-audio

[30] https://sail.usc.edu/iemocap/iemocap_release.htm

[31] https://www.kaggle.com/datasets/debarshichanda/goemotions

- **Construction of the Proposed MVRS Dataset**
  - ❖ **Dataset Details**

The proposed multimodal dataset consists of eye tracking, body motion, GSR, and EMG data from 13 participants aged 12 to 60 (both genders), who were stimulated by VR emotional videos for five emotions: relaxation, fear, stress, sadness, and joy. The demographic data of the dataset is limited to Iran, as the recording took place in Tehran, Iran. The recording environment was a simple room in a residential house, and the whole recording lasted less than a week in April 2025. Data collection for all modalities contains timestamps for precise synchronization in the multimodal fusion stage. Figure 5 depicts Russell's circumplex model of emotion [64], which is modified a bit in order to represent our collected emotions. Russell's circumplex model of emotion organizes emotions in a circular structure based on two dimensions: valence (ranging from pleasant to unpleasant) and arousal (ranging from high to low activation). This model shows how emotions blend smoothly into one another rather than existing as isolated categories. In this model, joy lies in the high-arousal, high-valence quadrant, representing a pleasant and energetic state. Sadness, in contrast, falls in the low-arousal, low-valence area, reflecting a deactivated and unpleasant feeling. Fear is positioned in the high-arousal, low-valence zone—it's unpleasant but highly activating, often linked with urgency or threat. Stress also resides near fear but may cover a broader range depending on intensity, typically characterized by moderate to high arousal and low valence. These emotions demonstrate how arousal and valence together define emotional experiences and how closely related states like stress and fear differ primarily in arousal intensity or situational context.

---

[32] https://github.com/sarnthil/unify-emotion-datasets/tree/master/datasets
[33] https://www.kaggle.com/datasets/jackksoncsie/emobank
[34] https://aclanthology.org/I17-1099/
[35] https://github.com/emorynlp/emotion-detection

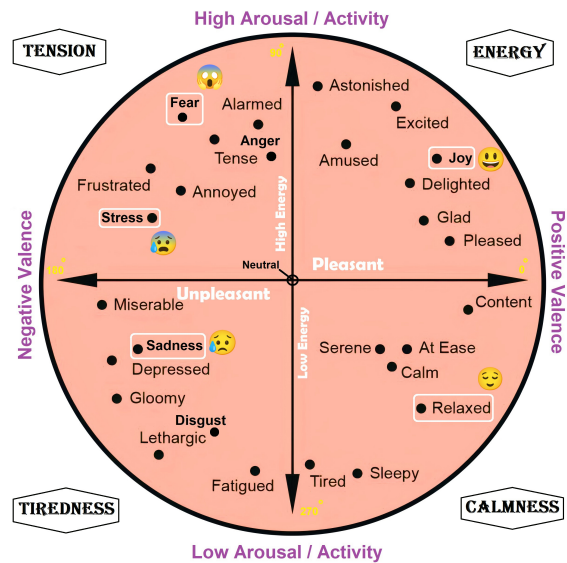

Figure 5. A bit modified Russell's circumplex model of emotion based on our collected emotional data

❖ **VR Stimuli Videos**

A rational order is considered for the stimulus videos. The video starts with a countdown, accompanied by a beep sound, which alerts the participant. This is followed by the first stimulus video, a short starter relaxation video that lasts for 20 seconds. The video is a beautiful, constant scene of a rainfall inside a cave covered with trees and plants, as shown in Figure 6. This video is taken from Stanford University's Virtual Human Interaction Lab website[36]. This data is essential for comparison with stress responses and for developing relaxation training, biofeedback, and wellness monitoring systems. Relaxation place at the bottom right corner of the 2D emotional model, having a high level of pleasantness and a low level of energy.

The next video is our first fear emotion stimulus, which lasts 1 minute and 10 seconds, and is a scene from the Evil Within 2 (2017) video game[37]. It is a selected scene from Laura Encounter, which features a creepy female creature. During the transition from relaxation to fear, activation of the amygdala and the Hypothalamic Pituitary Adrenal (HPA) axis leads to the release of cortisol and adrenaline, engaging the sympathetic nervous system. This results in physiological changes such as pupil dilation, accelerated saccadic eye movements, and reduced fixation durations, reflecting increased visual vigilance. The GSR rises due to enhanced sweat gland activity, indicating heightened arousal. Simultaneously, EMG signals from the hand reveal increased muscle tension and micro-movements associated with stress reactivity. These multimodal responses collectively mark a shift to a state of elevated emotional and physiological alertness. The fear stimulus continues with another video clip, which lasts for 1 minute and 52 seconds from The Thing (2011) movie[38]. It is a horrifying scene in which a human converts into an alien and runs after the main character. In both fear video clips, the main character survives the situation in which brings some relief to the participants and makes them ready for the next emotional clip. Fear was elicited to capture acute autonomic and behavioral responses to perceived threat, characterized by rapid physiological shifts. The collected data support applications in Post-Traumatic Stress Disorder (PTSD) assessment, anxiety research, phobia desensitization, adaptive VR therapies, emotion-aware HCI, and early mental health diagnostics.

In total, we have four relaxation videos that take place at the beginning of the experiment and in between other emotions. Right after the fear elicitation video, we have another relaxation video from the shallow reef zone with a beautiful sea blue background for 20 seconds, the same Stanford repository as for the first relaxation video. Short relaxation videos were inserted between emotional clips to reset the participant's physiological state and reduce emotional carryover. This makes cleaner, independent emotional responses and minimizes habituation or overlap. It also helps maintain participant comfort and emotional balance during extended VR exposure. Relaxation reduces amygdala activity and lowers cortisol, shifting the body toward parasympathetic dominance. Eye movements are slow, with longer fixations and reduced saccades. GSR levels drop due to decreased sweat gland activity. EMG shows minimal muscle tension, and body motion stabilizes with fewer micro-movements.

---

[36] https://vhil.stanford.edu/

[37] https://www.polygon.com/2017/10/12/16461738/the-evil-within-2-review

[38] https://www.imdb.com/title/tt0905372/

After the fear stimulus, the participant will be elicited with another emotion from the top left quarter of the 2D emotional model, called stress. The stress is similar to fear in arousal to high value, but not in valence, as it is low valence in the opposite of fear, which has high valence. The first VR stress video clip lasts for 2 minutes and 44 seconds from the Saw 2 movie[39] (2005). The character in the scene finds himself in a situation where he has to save his life from an evil trick. His neck is tied up to a trap that will end his life at a specific time. He has that specific time to extract the key from the back of his eye with a scalpel; the whole situation and ticking clock elicit a high level of stress in the participant. Stress activates the HPA axis just like fear, increasing cortisol and sustaining sympathetic nervous system activity. Eye tracking shows heightened vigilance with faster saccades and shorter fixations. GSR levels rise steadily, reflecting prolonged arousal and increased sweat gland activity. EMG, on the hand, captures persistent muscle tension and possible tremors. Body motion becomes less fluid, showing subtle rigidity or restless micro-adjustments linked to ongoing strain. Stress was elicited to capture sustained physiological and behavioral responses to prolonged cognitive or emotional pressure, marked by gradual but persistent arousal. The collected data supports applications in chronic stress monitoring, burnout detection, workload analysis, adaptive biofeedback systems, emotion-aware HCI, and early intervention in stress-related mental health conditions. The stress elicitation goes to the next stress stimulus video directly, which is a roller coaster experience that lasts for exactly 1 minute. The video is taken from Instagram[40]. The roller coaster videos are proven to have a high level of stress elicitation [65].

Right after a stress stimulus, a relaxation video from the Stanford repository will be played, lasting exactly 20 seconds, similar to the second relaxation video but with a lot of divers in it. This resets participants' emotions to neutrality for the next emotion, which is sadness.

The next stimulus video clip belongs to the sadness emotion, lasting 1 minute and 46 seconds from the A Quiet Place movie[41] (2018). This emotion belongs to the bottom left quarter of the emotional model in Figure 5, having low arousal and low valence. The selected scene depicts the father of the family sacrificing himself to save his children, and so by letting the creature get him, in that case, his children have time to escape. Sadness engages brain regions such as the anterior cingulate cortex and subgenual prefrontal cortex, often linked with emotional regulation and withdrawal. It may suppress sympathetic activity over time, leading to decreased arousal and a shift toward parasympathetic dominance. Eye tracking reveals reduced saccadic activity, longer fixation durations, and lowered visual engagement. GSR levels show slight increases or flat responses, depending on emotional intensity. EMG from the hand reflects low muscle tension or slight tremors linked to emotional heaviness. Body motion becomes minimal, often showing reduced movement or slumped posture associated with decreased energy and motivation. Sadness was elicited to examine the psychophysiological signatures of low-arousal negative affect, marked by emotional withdrawal and reduced engagement. The collected data support applications in depression screening, affective computing, mood monitoring in immersive environments, and the development of empathetic human-computer interfaces for emotional support and intervention.

We have another beautiful relaxation scene for 20 seconds from the same Stanford repository from the sea bottom, to stabilize emotions for the last emotion of joy. The joy emotion is associated with high arousal and high valence and meets the top right quarter of the emotional model. The joy stimulus video lasts for 1 minute and 47 seconds from Mr. Bean[42]. It is a very funny scene in which Mr. Bean goes to the doctor, and a lot of funny scenarios happen. It is needed to end the experiment with happiness, as they have to forget about sad, fearful, and stressful scenes psychologically. Joy activates key brain regions such as the ventral striatum, nucleus accumbens, and medial prefrontal cortex, associated with reward processing, positive affect, and motivation. It enhances parasympathetic activity while modulating dopaminergic and oxytocin-related pathways, promoting relaxation and social bonding. Eye tracking typically shows smooth, engaged tracking patterns with longer fixations on positive stimuli. GSR responses will exhibit moderate, short bursts of arousal due to emotional excitement. EMG signals from the hand show reduced muscle tension with possible micro-expressions of physical ease or subtle, relaxed movements. Body motion becomes more fluid or expressive, indicating openness and high energy. Joy was elicited to observe the physiological and behavioral markers of high-arousal positive affect, reflecting reward, engagement, and emotional uplift. The collected data supports applications in positive emotion modeling, well-being monitoring, human-robot interaction, immersive therapy design, and adaptive systems that respond to user mood and engagement. The total VR video elicitation experience will last for 11 minutes and 39 seconds. Also, we employed VaR's VR Video Player app[43] as the VR player of our stimulus videos. Snapshots and the order of the stimulus videos are visible in Figure 6. Figure 7 illustrates the effect of each emotion on different parts of the human brain.

---

[39] https://www.imdb.com/title/tt0432348/
[40] https://www.instagram.com/reel/DB7scL3pZg2/
[41] https://www.imdb.com/title/tt6644200/
[42] https://www.youtube.com/watch?v=2v3mLxd2FfA

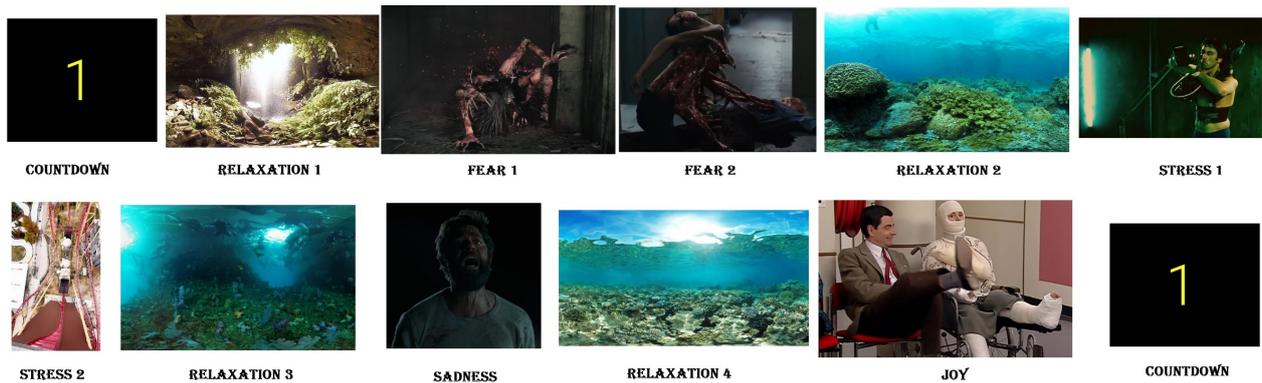

Figure 6. Snapshots and the order of our stimulus VR videos during the data collection

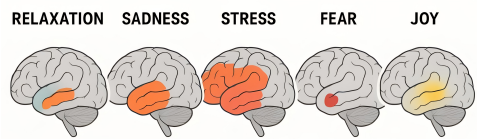

Figure 7. The relation between each emotion and different parts of the human brain

❖ **VR HMD Setup**

Figure 8 illustrates the complete setup of our custom-built eye-tracking system, starting from left to right. The core of the system is a VR SHINECON HMD[44], visible in the VR HMD front view. This standard VR headset has been carefully modified manually by us to accommodate an FHD USB webcam (30 fps) for real-time eye tracking. In the VR HMD Inside View, we can see the internal placement of the webcam positioned to face the user's eye, providing accurate capture of eye movements. To have proper illumination and enhance the visibility of the eye, especially in dim environments, a USB-powered LED light has been added inside the headset (around the webcam for the right eye), as shown in both the inside and VR HMD top view. The LED light is carefully covered with some white paper tape in order to prevent hurting the participant's eye. The LED light and the webcam are wired externally through the headset, clearly visible in the top-view image, where the cabling is routed out of the HMD to connect to the processing unit. The MSI FHD USB webcam[45] is central to capturing high-resolution images of the eye for real-time analysis. The real-time eye-tracking experiment snapshot shows how the webcam feed is processed on a laptop, where the Spyder Python IDE[46] is used to track pupil movement dynamically. Finally, the last frame presents the full experimental setup in action, with a subject wearing the modified VR headset while being monitored in a controlled environment. This setup demonstrates a cost-effective and practical solution for real-time eye tracking using modified consumer-grade VR equipment. The VR HMD is a basic model that accepts a smartphone, in which we used a Motorola 6.5-inch smartphone for its display. Due to a low budget, we just modified a simple VR headset so eye tracking data is acquired by the right eye, and the participant watches the VR video from the left eye. Also, we used two laptops for the whole experiment, in which the system with lower specifications (Core i5 CPU with onboard graphics card) was dedicated to the eye tracking task, and the stronger one (Core i7 CPU with RTX 4060 graphic card) to the body motion and physiological data collection tasks. We collected timestamps and their corresponding frame numbers, as well as the MP4 format, using the webcam for the final multimodal fusion step. We will synchronize each modality's timing using their timestamps, so we know what emotional response happened for each modality at a time. In that case, we can detect the power and weakness of each modality corresponding to each emotion. However, it depends on the participant's states, such as age, gender, ethnicity, culture, illnesses, current mental engagements, and any type of food or medicine consumption before the experiment, too.

---

[43] https://play.google.com/store/apps/details?id=com.abg.VRVideoPlayer&hl=en
[44] https://www.amazon.de/VR-SHINECON-Android-Handy-Virtual-Reality-3D-Brille-SC-G05A-white/dp/B0CYWZ7Q5C
[45] https://www.amazon.de/MSI-Webcam-PROCAM-FHD-H01-0001855-Mehrfarbig/dp/B0BBBYH7X2
[46] https://www.spyder-ide.org/

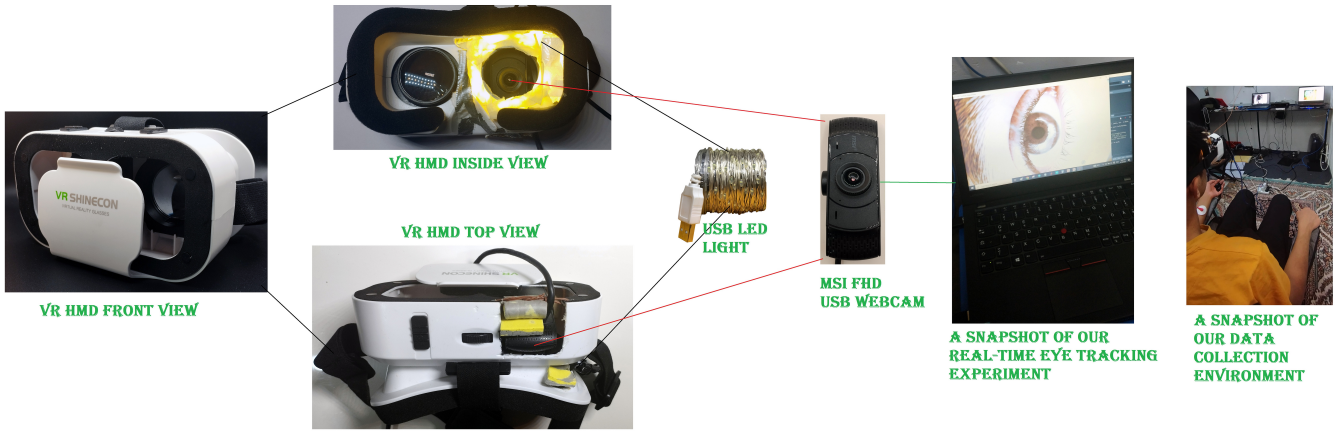

Figure 8. Our modified VR HMD for the data collection

❖ **Kinect Setup**

Figure 9 presents our body motion tracking setup using Microsoft Kinect v2, shown in a sequence from left to right. At the center of the setup is the Microsoft Kinect v2 sensor, highlighted in the close-up image. Our Kinect's cooling fan, which was at the back of the sensor, was broken, so we installed an external 12-Volt cooling fan from the side, as shown in Figure 9. This depth-sensing device captures full-body skeletal movements in real time. On the left, you can see the Kinect connected to a laptop running body tracking, displaying depth-based skeletal data. The image on the right shows a participant seated while wearing the VR headset, engaged in the experiment. The Kinect sensor is placed in front of the participant, capturing body movements, which are then visualized on the laptop screen, as shown in the snapshot of the experiment. This setup allows for simultaneous VR-based stimulation and body motion tracking, making it suitable for multimodal HCI and behavioral analysis experiments. The Microsoft Kinect v2, originally developed for the Xbox One and later adopted for research and development applications, is a powerful depth-sensing and motion-tracking device. It comprises three main sensing modules: an RGB camera, an infrared (IR) emitter with a time-of-flight depth sensor, and a multi-array microphone system. Unlike its predecessor, which used structured light for depth estimation, Kinect v2 uses time-of-flight technology to capture high-resolution depth data at 512×424 pixels and up to 30 frames per second. The RGB camera delivers 1080p color video, while the IR sensor operates independently to calculate the time light takes to bounce back from objects, allowing for accurate 3D reconstruction of the environment. One of Kinect v2's most notable features is its skeletal tracking capability, which can detect and track up to six human bodies simultaneously and recognize 25 individual joints per person. This makes it exceptionally well-suited for body motion capture applications in areas like healthcare, sports science, rehabilitation, animation, and interactive systems. The device also includes an active cooling system, a USB 3.0 interface, and requires the Kinect for Windows SDK 2.0 for development. Its ability to deliver real-time joint positions and body orientation data with high spatial and temporal accuracy has established it as a popular tool for non-invasive motion analysis, gesture recognition, and human-computer interaction research. We used the Kinect Studio software from the Kinect SDK,[47] A tool to capture in eXtensible Event Format (XEF) format. It is a rich data type that will be explained later in the paper. Kinect collects following body joints: SpineBase (0), SpineMid (1), Neck (2), Head (3), ShoulderLeft (4), ElbowLeft (5), WristLeft (6), HandLeft (7), ShoulderRight (8), ElbowRight (9), WristRight (10), HandRight (11), HipLeft (12), KneeLeft (13), AnkleLeft (14), FootLeft (15), HipRight (16), KneeRight (17), AnkleRight (18), FootRight (19), SpineShoulder (20), HandTipLeft (21), ThumbLeft (22), HandTipRight (23), and ThumbRight (24). The data is collected with the stronger system, as mentioned earlier, alongside physiological signals.

---

[47] https://www.microsoft.com/en-us/download/details.aspx?id=44561

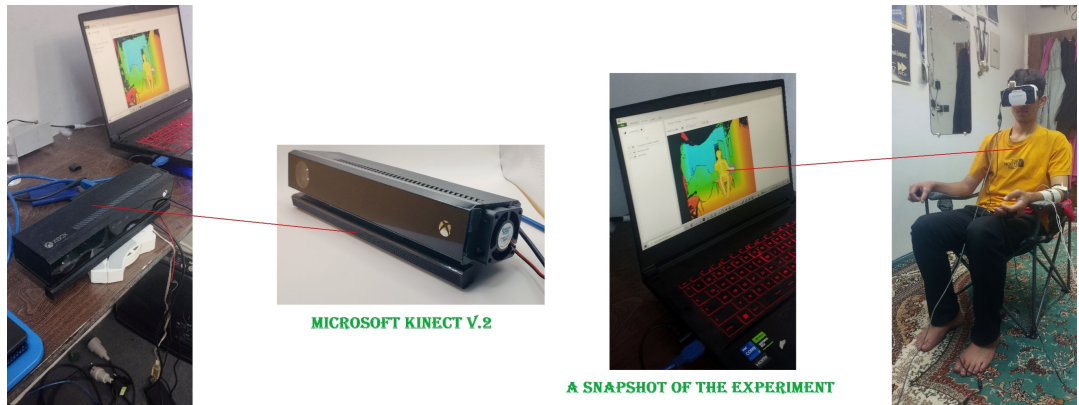

Figure 9. Our Microsoft Kinect V.2 setup for body motion modality data collection

❖ **Arduino Setup**

Figure 10 depicts a detailed and structured setup designed for emotion recognition through physiological signal acquisition using both EMG and GSR sensors installed on an Arduino UNO board. The Arduino Uno board[48] is the central microcontroller used in our emotion recognition setup, serving as the primary interface between the physiological sensors and the data processing environment. Based on the ATmega328P microcontroller, the Arduino Uno offers a reliable and accessible platform for analog signal acquisition, making it ideal for real-time biosignal monitoring. The board is powered via USB and also supports external battery input, making stable operation during extended sessions. It communicates with the host computer through a USB serial interface. With its 10-bit ADC resolution, straightforward pin layout, and vast software support, the Arduino Uno enables seamless integration of hardware and software in this physiological data acquisition system, playing a pivotal role in capturing the emotional responses induced by VR stimuli. The EMG sensor module[49] used in the setup measures the electrical activity produced by skeletal muscles. The module features a 3.5mm electrode input jack for connecting surface EMG electrodes, which are typically placed on the skin over targeted muscles. It amplifies and filters the raw biopotential signals using onboard components such as an instrumentation amplifier and several passive elements. The processed signal is then sent through the SIG (signal) pin, while the +VS and GND pins supply power. The output analog signal is read by the Arduino Uno through analog pin A0, providing real-time muscle activity monitoring. This sensor plays a crucial role in our emotion recognition system by capturing muscle tension changes in response to VR-induced stimuli. The GSR module[50] is designed to measure the electrical conductance of the skin, a parameter that varies with its moisture level and is closely related to emotional arousal. The module consists of a signal conditioning circuit and a connector for electrode pads that attach to the fingers or palm. It provides an analog voltage output through the SIG pin, which is read by the Arduino via analog pin A1 in the setup. The other pins include VCC for power (3.3V or 5V), GND for ground, and NC (not connected). This compact module is used to capture subtle physiological changes during emotional reactions, making it ideal for applications like stress detection, biofeedback, and emotion recognition in VR experiments.

Based on Figure 10, the core of the system is an Arduino Uno board (highlighted in blue), which serves as the data acquisition hub. The EMG sensor (marked in green) is connected to analog pin A0 of the Arduino and is applied to the subject's forearm using medical-grade electrodes to measure muscle activity in response to emotional stimuli. Similarly, the GSR sensor (marked in red), responsible for detecting skin conductivity changes related to arousal and stress, is connected to analog pin A1. The EMG sensor is powered using two 9-volt batteries (highlighted in orange), making it portable and providing stable power delivery during data collection. The subject wears a VR headset, driven by a smartphone screen (highlighted in purple), which presents emotionally charged video content. The controller (highlighted in yellow) is used only once at the beginning of each session to trigger the video playback. Simultaneous data recording is achieved using Arduino IDE[51] for serial communication and Spyder IDE (Python environment) for real-time logging and processing, both running synchronously. The Arduino communicates over COM3 at a baud rate of 115200 and with a recording delay of 0.2 seconds, enabling high-frequency signal acquisition without data loss. The entire setup is carefully designed to synchronize VR-based emotional stimuli with biosignal capture, providing high-quality multimodal data for emotion recognition research. A wiring diagram and real-time photos of the system components further validate and visualize the hardware configuration and its use in practice.

---

[48] https://store.arduino.cc/products/arduino-uno-rev3?queryID=3cf9899d39f9f98614f6a1f5bd7b776a
[49] https://robu.in/product/advancer-technologies-emg-muscle-sensor-v3-0-with-cable-and-electrodes/
[50] https://wiki.seeedstudio.com/Grove-GSR_Sensor/
[51] https://www.arduino.cc/en/software/

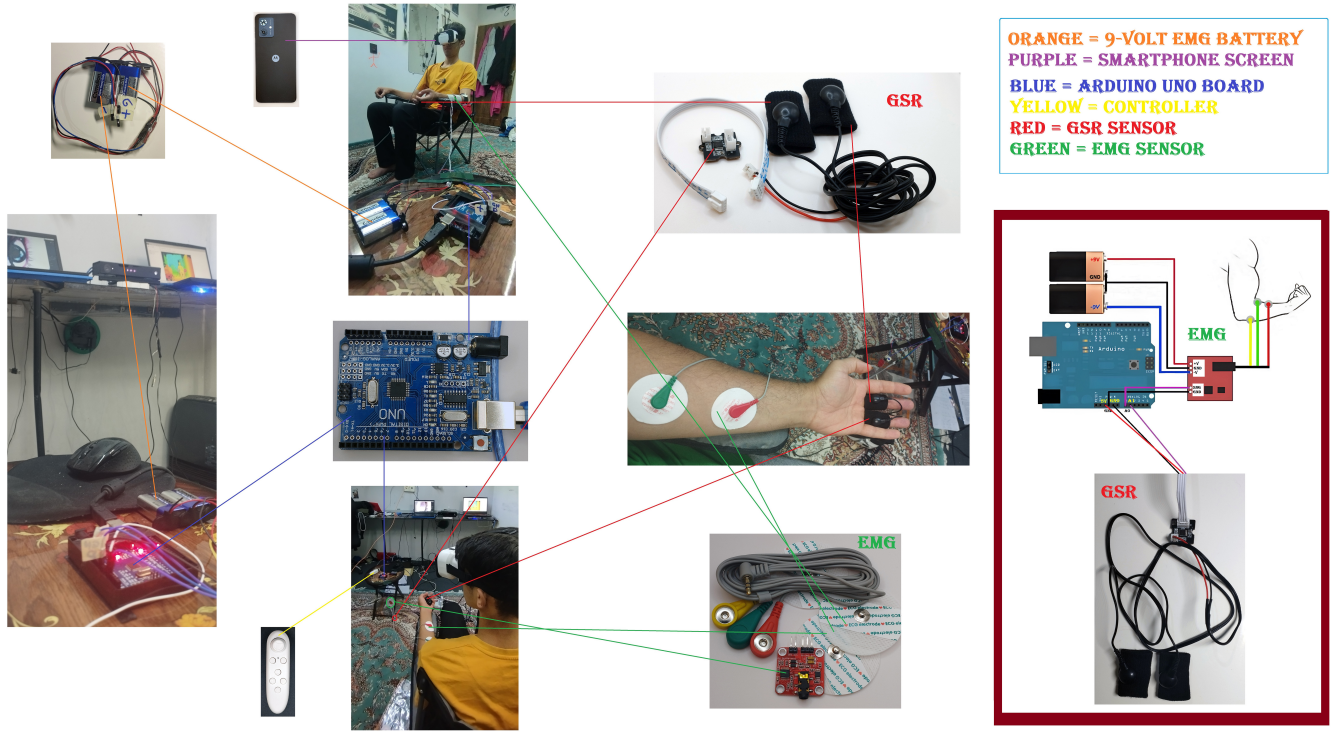

Figure 10. Our Arduino setup for EMG and GSR data collection

❖ **Data Types**

For eye tracking, the FHD MPEG-4 Part 14 or MP4 with 30 fps is recorded, and simultaneously, TeXT (TXT), Comma Separated Values (CSV), and Excel Spreadsheet (XLS) files having the timestamp and video frame number are stored, too. The MP4 can store video, audio, subtitles, and still images, making it ideal for streaming and sharing across platforms. MP4 is favored for its high compression efficiency and broad compatibility with devices and media players. The TXT extension represents a plain text file that contains unformatted text. It is one of the simplest and most universal file formats, readable by almost any text editor or word processor. TXT files are commonly used for notes, logs, configuration data, and documentation, as they store only raw text without fonts, colors, or other rich formatting. The CSV is used to store tabular data in plain text format. Each line in a CSV file represents a data row, and values are separated by commas (or sometimes other delimiters like semicolons). It is widely used for data exchange between applications like Excel, databases, and programming environments because of its simplicity and compatibility. The XLS is used for Microsoft Excel spreadsheet files created in versions before Excel 2007. It stores data in a binary format and can include complex features like formulas, charts, pivot tables, and macros. XLS files are commonly used for organizing, analyzing, and visualizing numerical and tabular data. While newer Excel versions use the .xlsx format, XLS remains compatible with many older systems and applications.

For body motion, and as mentioned above, the extension is eXtensible Event Format (XEF) in which is recorded by the Kinect studio software. The XEF is a proprietary file format developed by Microsoft for recording Kinect sensor data, primarily used within Kinect Studio. It captures a complete, time-synchronized stream of all available Kinect outputs, including RGB color video, depth maps, infrared frames, audio input, and most importantly, detailed body motion data. The body motion stream contains 3D joint positions, orientations, tracking states, and skeletal structures for up to six individuals, making XEF crucial for applications involving gesture recognition, rehabilitation, biomechanics, and motion analysis. We extracted the body motion part from XEF files, which will be explained in the data processing section.

As for EMG and GSR data, we stored timestamps and mentioned physiological signal values with a 0.2-second delay in TXT, CSV, and JSON formats. The JavaScript Object Notation (JSON) is a lightweight, text-based data format used for storing and exchanging structured information between systems. It represents data as key-value pairs, arrays, and nested objects, making it both human-readable and easy for machines to parse. Widely used in web development and APIs, JSON is language-independent but follows conventions familiar to programmers of the C family, such as JavaScript, Python, and Java. Its simplicity and flexibility have made it a standard format for data interchange across various platforms and applications.

❖ **Data Collection Protocol**

Our data collection protocol, as shown in Figure 11, is very simple and follows almost the summary base of all data collection protocols for such tasks by other researchers from the previous decade available on the web. Before starting the data collection for the Multimodal Virtual Reality Stimuli-based emotion recognition dataset (MVRS), each participant will first be approached and given a clear explanation of the study's goals, procedures, and the types of data to be collected, including EMG, GSR, eye tracking, and depth images for body motion analysis. Participants will be informed that they will watch emotionally evocative video clips in a VR environment and that their physiological and behavioral responses will be recorded non-invasively. We will emphasize that their identity will remain confidential, as no facial, vocal, or fingerprint data will be stored. After this explanation, participants will be asked to review and complete a questionnaire regarding their basic information and health status to ensure eligibility. They will then be presented with a detailed consent form, and only those who voluntarily agree to participate by signing the form will proceed. Upon receiving their written consent, we will begin the data recording session under safe and comfortable conditions, ensuring participants can withdraw at any point without penalty. It has to be mentioned that the final data is recorded after multiple pilot tests.

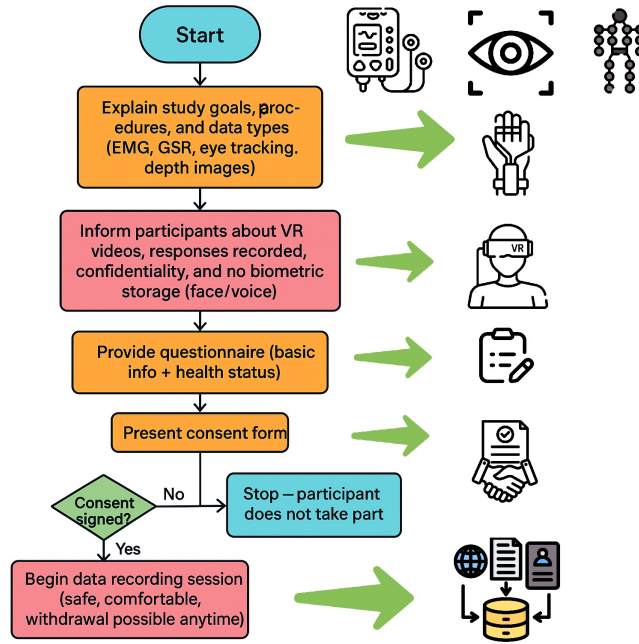

Figure 11. Our data collection protocol for the MVRS dataset

❖ **Challenges**

During the data collection phase, several challenges emerged that required technical adaptation and creative problem-solving. One of the main issues was configuring the Arduino board to simultaneously handle EMG and GSR signals through separate analog pins, which involved careful calibration and stability testing. Running both the Arduino IDE for live sensor monitoring and Spyder for real-time data logging on the same machine proved difficult, requiring synchronization and port management. Another significant challenge was identifying appropriate emotional video stimuli that could elicit strong, diverse emotional responses across different age groups and genders without being culturally or personally offensive. Additionally, recording all modalities, including physiological signals, eye tracking, and body motion, using two separate laptops introduced synchronization issues, especially with ensuring consistent timestamps across devices. Extracting body motion data from Kinect's .xef files was another technical hurdle, demanding the use of specialized SDKs and scripts to accurately convert raw depth and skeletal data into usable joint coordinates. A particularly complex task was modifying a basic VR headset to incorporate an internal webcam for eye tracking; this required physical reconstruction of the headset, secure mounting of the camera inside, and integrating an illumination system to ensure the eye was captured clearly without affecting the VR experience. Each of these challenges played a crucial role in shaping a robust and reliable multimodal data collection setup.

- **Evaluation of the MVRS Dataset**
  ❖ **Data Processing**

Each participant's timestamp for each modality is carefully recoded for data cleansing and the modality synchronization step. So, the first step is to extract exact data and remove extra collected data. However, the data is recorded simultaneously, but some data gets some extra samples due to technical and hardware issues; the timestamps are recorded, which fixes the problem. For body motion modality, there is an extra step, as it is not text or spreadsheet-based output. It is Xef data that opens by Kinect studio, so we used a C# code to extract each body joint's data from the depth-based video files. The second step is labeling based on the time stamps. As we have the exact duration of each stimulus and also the exact duration of each participant's experiment, we labeled the data for each row of modalities' raw data accordingly. The third step is to remove and replace any Nan cells depending on the modality. The fourth and last step is to stack all participants' data into a single file for each modality. In that case, we can perform feature extraction, classification, and data analysis on them separately. Additionally, we fused extracted features of all modalities using different approaches for the same processing, which is explained in the next subsections.

❖ **Feature Extraction**

The data is processed and cleaned by the previews stage, and it is ready for further processing. The raw data is full of redundant and even repetitive samples, leading to unreliable results and overfitting. To get some meaning out of the data, for each modality, specific and related features are extracted. These features are extracted based on each participant and each emotion label separately in order to have the full effect.

For the body motion modality, we extracted a comprehensive set of kinematic, dynamic, and spectral features from the skeletal joint data to capture fine-grained patterns associated with emotional expression. For each joint's 3D position and rotation, we computed statistical descriptors such as mean, standard deviation, range, root mean square, and entropy to characterize overall motion variability and distribution. To capture dynamic aspects, we derived velocity, acceleration, and jerk profiles along each axis, along with peak-related measures such as the number of motion peaks and their temporal intervals, reflecting the rhythmicity and abruptness of movements. Furthermore, we incorporated spectral-domain features, including dominant frequency and spectral energy, to quantify periodicity and motion intensity in the frequency space. These features provided a rich representation of both spatial and temporal characteristics of body movements, enabling the identification of subtle behavioral cues that differentiate emotional states [66, 67].

For the physiological modality, we extracted a comprehensive set of time-domain and frequency-domain features from both GSR and EMG signals to capture autonomic and muscular dynamics associated with emotional states. From GSR, we derived statistical measures such as mean, standard deviation, minimum, maximum, and Area Under the Curve (AUC) to quantify overall signal magnitude and variability. Additionally, we computed peak-based descriptors including the number of skin conductance responses, their mean amplitude, rise times, and peaks per second, reflecting phasic arousal activity linked to sympathetic nervous system responses. For EMG, we extracted features characterizing muscle activation intensity and frequency composition, including mean, RMS, Mean Absolute Value (MAV), Zero-Crossings (ZC), Slope Sign Changes (SSC), Waveform Length (WL), and Integrated EMG (IEMG), which capture both fine-grained temporal fluctuations and muscle contraction complexity. Furthermore, we used spectral measures such as the mean frequency and median frequency obtained via Welch's power spectral density, allowing us to assess muscle fatigue and activation patterns relevant to emotional expression. Together, these features provided a rich representation of physiological responses, making a more accurate characterization of arousal and valence dimensions crucial for multimodal emotion recognition [68-70].

For the eye-tracking modality, we extracted a comprehensive set of pupil, iris, eyelid, and gaze-related features from video recordings to capture oculomotor dynamics associated with emotional responses. From the pupil, we derived geometric descriptors including diameter, area, aspect ratio, circularity, eccentricity, and rotation angle, which quantify dilation patterns and shape changes reflecting cognitive and emotional arousal. We also estimated iris color distributions (RGB channels) to capture subtle variations in appearance under different lighting or physiological states. Eyelid openness was computed using image projection statistics, providing an indicator of fatigue, alertness, and expressive behavior. Additionally, entropy-based measures were calculated from localized intensity distributions around the pupil to capture texture variability linked to emotional engagement. Dynamic gaze characteristics, including pupil velocity along x/y axes, fixation stability, and time since last blink, were also extracted to represent attentional control and emotional reactivity. Together, these features provide a high-resolution representation of ocular behavior, making the detection of subtle psychophysiological responses that play a critical role in differentiating emotional states [71-73].

❖ **Multimodal Fusion**

Multimodal fusion has been shown as a powerful strategy for enhancing emotion recognition by integrating complementary information from diverse data sources such as body motion, physiological signals (for instance,

EMG, GSR), and eye-tracking features. Emotions are inherently complex, involving intertwined physiological, behavioral, and cognitive processes; thus, relying on a single modality often leads to incomplete or ambiguous representations of emotional states. By combining modalities at different fusion levels, early (feature-level), late (decision-level), or hybrid approaches, multimodal frameworks use cross-modal correlations and capture richer representations of affective dynamics. In emotion recognition, fusion improves robustness against noise, variability across participants, and missing data, while providing a more holistic understanding of emotional expressions. For example, body motion encodes expressive gestures, physiological responses reflect autonomic arousal, and eye-tracking reveals attention and cognitive engagement; their integration enables systems to detect subtle patterns that may be invisible within a single modality. Recent studies have demonstrated significant performance gains through multimodal fusion, getting higher classification accuracy, better generalization, and improved interpretability in both controlled and real-world environments. Consequently, multimodal fusion represents a critical advancement for building reliable, adaptive, and human-centered emotion recognition systems [74-76]. Here, we used three multimodal fusion approaches. Two from the early stage and one from the late stage.

Early-stage fusion combines raw or preprocessed features from multiple modalities into a single unified feature space before feeding them into a learning model. This approach gets cross-modal correlations directly at the feature level, enabling the system to learn richer joint representations but requiring careful alignment and feature selection to handle differences in scale, sampling rates, and dimensionality. Late-stage fusion, on the other hand, processes each modality independently and combines its outputs or decisions at the final stage, often using strategies like majority voting, weighted averaging, or ensemble methods. This approach preserves modality-specific characteristics, is more robust to noisy or missing data, and provides greater flexibility, though it may capture fewer inter-modal dependencies compared to early fusion.

For the first fusion strategy, we employed an early-stage multimodal fusion approach to integrate body motion, EMG+GSR, and eye-tracking features into a unified representation space for emotion recognition. Since the three modalities had different temporal resolutions and sampling rates, we first applied interpolation-based temporal alignment to resample all feature sets to a common length, confirming synchronized data across modalities. To address the high dimensionality and reduce redundancy, we performed modality-specific feature selection: mutual information-based selection [77] for body motion, ANOVA F-statistic-based ranking [78] for EMG+GSR, and variance thresholding [79] for eye-tracking, retaining only the most discriminative and stable features within each modality. After selection, we concatenated the optimized feature subsets from all modalities into a single fused feature space, resulting in a compact yet information-rich representation of multimodal emotional cues. This early fusion design enabled the model to exploit cross-modal correlations and learn integrated emotional patterns, improving the ability to capture subtle behavioral, physiological, and ocular dynamics simultaneously.

For the second early-stage fusion strategy, we adopted a feature-level integration combined with dimensionality reduction via an autoencoder [80] to create a compact yet discriminative representation of the multimodal dataset. Initially, body motion, EMG+GSR, and eye-tracking features were temporally aligned through interpolation and concatenated into a unified feature space. Since the raw feature set was high-dimensional and potentially redundant, we standardized all features and trained a fully connected deep autoencoder to learn compressed latent representations while preserving essential information [81]. The encoder projected the concatenated multimodal features into a lower-dimensional subspace (up to 100 dimensions or fewer), capturing cross-modal dependencies and complex nonlinear relationships between modalities. After encoding, we applied variance thresholding to remove near-constant features, further refining the feature space and improving efficiency. This approach used the strength of deep unsupervised representation learning to generate a compact, noise-robust embedding, enabling the model to decently exploit the complementary information present across modalities. By combining early fusion with learned feature compression, this method enhanced the generalizability and discriminative power of the multimodal emotion recognition framework while minimizing computational overhead.

For the third fusion strategy, we adopted a late-stage multimodal fusion framework, where the predictions from individual modalities, body motion, EMG+GSR, and eye-tracking, are combined at the decision level to exploit complementary information without forcing feature-level alignment. Initially, we performed interpolation to synchronize the number of samples across modalities and applied dimensionality reduction using Principal Component Analysis (PCA) within each modality independently, retaining the most informative components while mitigating redundancy and noise. Unlike early fusion approaches, this method preserved the unique statistical properties of each modality and allowed independent optimization of their processing pipelines. After feature transformation, predictions from the separate modalities were aggregated using a majority-voting mechanism, where the final outcome is determined by the consensus across modalities. Majority voting is a late-stage fusion technique where the final decision is based on the most common prediction (votes) among multiple models or modalities. Each modality outputs its predicted class, and the class receiving the highest number of votes is selected. For example, the body motion model predicts joy (1 vote), the EMG+GSR model predicts joy (1 vote), and the eye-tracking model predicts sadness (1 vote). Since joy gets 2 votes and sadness gets 1 vote, the final decision is joy. This strategy uses the diversity of modality-specific representations and improves robustness, especially when one modality may be noisy or partially unreliable. By integrating decision-level fusion with dimensionality reduction, this approach provides a flexible and interpretable framework for capturing cross-modal complementarities while maintaining computational efficiency and resilience to modality-specific variability [82, 83]. Feature extraction and multimodal fusion steps are summarized in Figure 12.

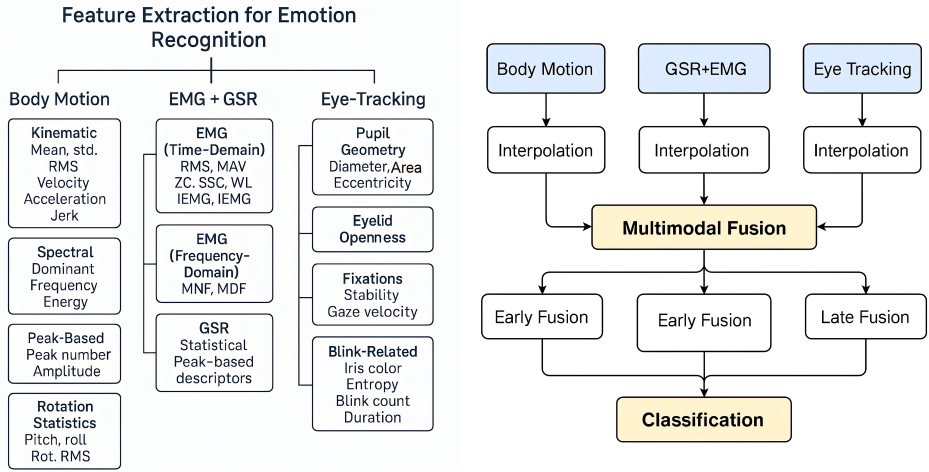

Figure 12. Summary of our feature extraction and multimodal fusion steps

❖ **Classification and Metrics**

In this study, we employed a comprehensive classification pipeline to evaluate the separated and fused modalities using a 5-fold cross-validation strategy. The pipeline processes the feature matrix by first removing the emotion and participant columns, followed by mean imputation to handle missing values. Subsequently, the feature space is standardized using z-score normalization and reduced via PCA, where the number of components is adaptively capped at 60 or the minimum of the number of training samples and available features to prevent overfitting. Three classifiers are evaluated independently: Random Forest (RF), Decision Tree (DT), and XGBoost (XGB), each trained (80 percent) and tested (20 percent) within the cross-validation folds.

RF combines multiple decision trees to improve accuracy and reduce overfitting. It is well-suited for multimodal emotion data due to its ability to handle high-dimensional features and capture complex relationships. DT creates simple, interpretable decision rules to classify data efficiently. They are useful for capturing non-linear patterns across modalities and serve as a strong baseline for performance comparison. XGB uses gradient boosting to sequentially improve tree-based models, offering high accuracy and robustness. It handles noisy, imbalanced multimodal features effectively, making it highly suitable for emotion classification [84].

For each fold, the model is fitted on the training set and tested on the held-out fold, and fold-wise classification accuracy is recorded. After completing all folds, the predictions are aggregated across splits to compute the final evaluation metrics. These include: (i) a complete classification report containing per-class precision, recall, and F1-score; (ii) the confusion matrix; (iii) seven one-vs-rest metrics per class, precision, recall (UAR), F1-score, one-vs-rest accuracy, one-vs-rest Cohen's kappa, one-vs-rest Matthews Correlation Coefficient (MCC), and support; and (iv) seven overall performance metrics, macro-averaged precision, macro-averaged recall/UAR, macro-averaged F1-score, overall accuracy, multiclass Cohen's kappa, multiclass MCC, and the mean cross-validation accuracy computed from the five folds.

Accuracy measures the proportion of correctly classified samples and provides an overall estimate of the model's effectiveness, but it can be misleading for imbalanced datasets. Precision quantifies the proportion of correctly predicted positive samples out of all positive predictions, making it important for minimizing false positives when distinguishing subtle emotional states. Recall (or Unweighted Average Recall) measures the proportion of correctly identified positive samples out of all actual positives, confirming that the model does not miss critical emotional cues. F1-score, the harmonic mean of precision and recall, provides a balanced measure when there is a trade-off between minimizing false positives and false negatives, which is especially useful for overlapping emotion classes. Cohen's Kappa evaluates the agreement between predicted and true labels beyond random chance, making it more robust than accuracy in cases of class imbalance. Finally, the Matthews Correlation Coefficient (MCC) offers a single balanced metric considering true and false positives and negatives, making it highly suitable for multi-class and imbalanced classification scenarios [85, 86]. All experiments are conducted consistently across the three classifiers using identical preprocessing, cross-validation splits, and evaluation criteria to have fair comparisons. Table 1 presents the classification performance for the body motion modality across five emotions using RF, DT, and XGB classifiers. Table 2 shows the evaluation results for the EMG and GSR physiological signals, which provide complementary information about emotional arousal and stress. Table 3 summarizes the performance of eye-tracking features, which reflect attentional patterns and visual engagement during emotional experiences.

Table 1. Evaluation Results for Body Motion Modality (5 folds – 20 % test)

| Class | Fear in % | | | Joy in % | | | Relaxation in % | | | Sadness in % | | | Stress in % | | | All Classes in % | | |
|---|---|---|---|---|---|---|---|---|---|---|---|---|---|---|---|---|---|---|
| Classifier/ Metric | RF | DT | XGB | RF | DT | XGB | RF | DT | XGB | RF | DT | XGB | RF | DT | XGB | RF | DT | XGB |
| Precision | 94.54 | 91.20 | 96.89 | 96.79 | 92.96 | 98.24 | 91.08 | 62.75 | 97.24 | 92.39 | 86.34 | 96.54 | 97.29 | 93.73 | 98.43 | 94.42 | 85.40 | 97.47 |
| Recall | 98.90 | 90.00 | 99.23 | 98.63 | 91.89 | 99.79 | 68.44 | 61.14 | 83.43 | 97.04 | 89.79 | 98.96 | 99.45 | 94.32 | 99.65 | 92.49 | 85.43 | 96.21 |
| F1-score | 96.67 | 90.60 | 98.05 | 97.70 | 92.42 | 99.01 | 78.15 | 61.94 | 89.81 | 94.66 | 88.03 | 97.74 | 98.36 | 94.02 | 99.04 | 93.11 | 85.40 | 96.73 |
| Accuracy | 98.62 | 96.21 | 99.20 | 99.02 | 96.81 | 99.58 | 95.67 | 91.51 | 97.86 | 98.35 | 96.32 | 99.31 | 98.93 | 96.14 | 99.38 | 95.30 | 88.49 | 97.66 |
| Cohen's Kappa | 95.80 | 88.22 | 97.54 | 97.08 | 90.40 | 98.74 | 75.81 | 57.16 | 88.62 | 93.69 | 85.86 | 97.33 | 97.56 | 91.18 | 98.57 | 93.91 | 85.15 | 96.97 |
| MCC | 95.84 | 88.23 | 97.55 | 97.09 | 90.41 | 98.74 | 76.76 | 57.15 | 88.95 | 93.73 | 85.89 | 97.34 | 97.57 | 91.18 | 98.58 | 93.97 | 85.16 | 96.99 |

Table 2. Evaluation Results for EMG+GSR Modality (5 folds – 20 % test)

| Class | Fear in % | | | Joy in % | | | Relaxation in % | | | Sadness in % | | | Stress in % | | | All Classes in % | | |
|---|---|---|---|---|---|---|---|---|---|---|---|---|---|---|---|---|---|---|
| Classifier/ Metric | RF | DT | XGB | RF | DT | XGB | RF | DT | XGB | RF | DT | XGB | RF | DT | XGB | RF | DT | XGB |
| Precision | 86.60 | 83.12 | 86.69 | 80.87 | 74.24 | 79.61 | 67.83 | 45.03 | 65.13 | 78.50 | 65.25 | 75.94 | 78.44 | 74.65 | 77.96 | 78.45 | 68.46 | 77.07 |
| Recall | 93.65 | 84.05 | 93.31 | 82.80 | 72.12 | 79.72 | 37.47 | 45.61 | 36.40 | 72.29 | 65.25 | 70.82 | 86.69 | 74.70 | 86.76 | 74.58 | 68.36 | 73.40 |
| F1-score | 89.99 | 83.58 | 89.88 | 81.82 | 73.19 | 79.66 | 48.28 | 45.32 | 46.70 | 75.27 | 65.25 | 73.29 | 82.36 | 74.68 | 82.12 | 75.54 | 68.40 | 74.33 |
| Accuracy | 94.56 | 91.38 | 94.51 | 94.11 | 91.53 | 93.48 | 91.60 | 88.49 | 91.31 | 92.74 | 89.38 | 92.12 | 88.06 | 83.72 | 87.86 | 80.54 | 72.25 | 79.64 |
| Cohen's Kappa | 86.26 | 77.74 | 86.12 | 78.13 | 68.17 | 75.78 | 44.12 | 38.89 | 42.38 | 71.02 | 58.98 | 68.68 | 73.37 | 62.68 | 72.97 | 74.33 | 63.88 | 73.15 |
| MCC | 86.38 | 77.74 | 86.23 | 78.32 | 68.18 | 75.78 | 46.43 | 38.88 | 44.52 | 71.11 | 58.98 | 68.73 | 73.58 | 62.68 | 73.21 | 74.54 | 63.89 | 73.35 |

Table 3. Evaluation Results for Eye Tracking Modality (5 folds – 20 % test)

| Class | Fear in % | | | Joy in % | | | Relaxation in % | | | Sadness in % | | | Stress in % | | | All Classes in % | | |
|---|---|---|---|---|---|---|---|---|---|---|---|---|---|---|---|---|---|---|
| Classifier/ Metric | RF | DT | XGB | RF | DT | XGB | RF | DT | XGB | RF | DT | XGB | RF | DT | XGB | RF | DT | XGB |
| Precision | 87.48 | 84.15 | 81.73 | 92.61 | 87.09 | 88.43 | 85.80 | 64.53 | 75.74 | 84.21 | 76.61 | 76.25 | 85.78 | 83.25 | 80.11 | 87.18 | 79.13 | 80.45 |
| Recall | 91.46 | 84.08 | 87.24 | 91.14 | 86.71 | 84.97 | 64.58 | 65.37 | 47.46 | 84.08 | 76.91 | 77.69 | 90.91 | 82.95 | 87.21 | 84.43 | 79.20 | 76.91 |
| F1-score | 89.43 | 84.12 | 84.39 | 91.87 | 86.90 | 86.66 | 73.69 | 64.94 | 58.35 | 84.14 | 76.76 | 76.97 | 88.27 | 83.10 | 83.51 | 85.48 | 79.16 | 77.98 |
| Accuracy | 94.38 | 91.74 | 91.61 | 97.51 | 95.97 | 95.97 | 94.73 | 91.93 | 92.26 | 95.20 | 92.95 | 92.96 | 92.27 | 89.20 | 88.98 | 87.04 | 80.90 | 80.89 |
| Cohen's Kappa | 85.60 | 78.54 | 78.67 | 90.40 | 84.52 | 84.29 | 70.83 | 60.39 | 54.33 | 81.32 | 72.60 | 72.81 | 82.51 | 75.17 | 75.26 | 83.06 | 75.21 | 74.91 |
| MCC | 85.61 | 78.53 | 78.66 | 90.42 | 84.55 | 84.26 | 70.88 | 60.88 | 54.30 | 81.30 | 72.57 | 72.77 | 82.50 | 75.10 | 75.32 | 83.13 | 75.21 | 75.07 |

Among the evaluated models in Table 1, XGB consistently outperforms RF and DT across nearly all metrics, getting the highest overall precision (97.47%), recall (96.21%), and F1-score (96.73%), showing highly reliable and stable predictions. High precision values (e.g., 98.43% for stress) show the model's ability to correctly identify positive cases with minimal false positives, while the strong recall scores (e.g., 99.65% for stress) confirm effective coverage of all true samples. The overall accuracy peaks at 97.66% with XGB, showcasing robust predictive performance across all folds. Moreover, the high Cohen's kappa (96.97%) and MCC (96.99%) indicate strong agreement between the predicted and true labels beyond chance, reinforcing the reliability of this modality. In contrast, DT exhibits weaker and more inconsistent performance, particularly for relaxation (precision = 62.75%, recall = 61.14%), suggesting that simple tree-based models may struggle to generalize the complex spatial-temporal patterns in motion data.

Compared to the body motion modality, the performance in Table 2 is more variable across classifiers and emotions, primarily due to the higher signal noise and individual differences in physiological responses. Among the classifiers, XGB again delivers the strongest overall performance, achieving the highest F1-score (74.33%), accuracy (79.64%), and MCC (73.35%), making it the most robust choice for this modality. The precision for most emotions remains moderate (e.g., 77.07% overall), meaning the model correctly predicts the majority of positive cases but occasionally misclassifies signals from overlapping emotional states. The recall is generally lower, particularly for relaxation (36.40% with XGB), indicating that subtle physiological changes associated with low-arousal states are harder to detect reliably. Cohen's kappa (73.15% for XGB) confirms a good but less perfect agreement compared to the motion modality, showing the inherent variability of physiological signals. DT performed the weakest overall, with lower scores in precision, recall, and MCC, proposing their limited ability to capture nonlinear relationships in the EMG and GSR feature space.

The results demonstrate moderate to strong classification performance across classifiers, with RF and XGB performing comparably well while DT lags behind slightly. XGB achieves the highest overall precision (80.45%) and F1-score (77.98%), while RF provides slightly better recall (84.43%), stating that RF captures more true emotional responses but at the cost of occasional false positives. The relatively high accuracy (up to 92.96% in some cases) shows that eye-based features are effective but slightly less discriminative than body motion patterns. Notably, stress and joy are classified more accurately across all classifiers, having a precision above 88%, while relaxation and neutral states are more challenging due to overlapping visual patterns and inter-participant variability. The Cohen's kappa and MCC values (~75% for XGB and RF) state moderate-to-strong agreement, supporting that eye dynamics provide valuable emotional cues but may require integration with other modalities for optimal recognition. Table 4 presents the results of the first early-stage fusion using feature selection techniques to combine body motion, EMG+GSR, and eye-tracking features. Also, Table 5 represents evaluation results for the second early-stage multimodal fusion method using the autoencoder. Finally, Table 6 shows the same achieved evaluation metric results for our employed late-stage fusion technique by majority voting.

Table 4. Evaluation Results for the First Early-Stage Fusion (Feature Selection Techniques - 5 folds – 20 % test)

| Class | Fear in % | | | Joy in % | | | Relaxation in % | | | Sadness in % | | | Stress in % | | | All Classes in % | | |
|---|---|---|---|---|---|---|---|---|---|---|---|---|---|---|---|---|---|---|
| Classifier/ Metric | RF | DT | XGB | RF | DT | XGB | RF | DT | XGB | RF | DT | XGB | RF | DT | XGB | RF | DT | XGB |
| Precision | 93.69 | 98.94 | 96.96 | 96.38 | 98.57 | 97.46 | 91.66 | 90.90 | 81.93 | 87.71 | 97.61 | 94.13 | 93.46 | 98.95 | 97.11 | 92.58 | 96.99 | 93.52 |
| Recall | 97.76 | 98.93 | 98.04 | 93.57 | 98.41 | 97.60 | 66.42 | 91.04 | 75.66 | 93.03 | 97.56 | 95.77 | 97.98 | 99.00 | 98.01 | 89.75 | 96.99 | 93.02 |
| F1-score | 95.68 | 98.94 | 97.49 | 94.96 | 98.49 | 97.53 | 77.03 | 90.97 | 78.67 | 90.29 | 97.58 | 94.95 | 95.67 | 98.97 | 97.56 | 90.72 | 96.99 | 93.24 |
| Accuracy | 97.71 | 99.45 | 98.69 | 98.47 | 99.53 | 99.24 | 95.47 | 97.93 | 95.31 | 96.97 | 99.27 | 98.46 | 97.16 | 99.34 | 98.43 | 92.89 | 97.76 | 95.06 |
| Cohen's Kappa | 94.12 | 98.56 | 96.61 | 94.05 | 98.21 | 97.08 | 74.58 | 89.80 | 76.04 | 88.50 | 97.15 | 94.03 | 93.55 | 98.49 | 96.40 | 90.70 | 97.10 | 93.58 |
| MCC | 94.10 | 98.55 | 96.54 | 94.09 | 98.21 | 97.00 | 73.99 | 90.01 | 76.03 | 88.52 | 97.10 | 94.05 | 93.51 | 98.48 | 96.22 | 90.79 | 97.10 | 93.59 |

Table 5. Evaluation Results for the Second Early-Stage Fusion (Autoencoder Dimensionality Reduction - 5 folds – 20 % test)

| Class | Fear in % | | | Joy in % | | | Relaxation in % | | | Sadness in % | | | Stress in % | | | All Classes in % | | |
|---|---|---|---|---|---|---|---|---|---|---|---|---|---|---|---|---|---|---|
| Classifier/ Metric | RF | DT | XGB | RF | DT | XGB | RF | DT | XGB | RF | DT | XGB | RF | DT | XGB | RF | DT | XGB |
| Precision | 97.30 | 97.63 | 97.19 | 98.09 | 97.19 | 97.91 | 92.96 | 83.26 | 87.69 | 94.29 | 94.99 | 93.84 | 97.58 | 97.79 | 97.54 | 96.04 | 94.17 | 94.83 |
| Recall | 99.06 | 97.64 | 98.48 | 97.92 | 97.12 | 97.78 | 81.52 | 83.68 | 78.88 | 97.24 | 94.85 | 96.62 | 99.07 | 97.72 | 98.68 | 94.96 | 94.20 | 94.09 |
| F1-score | 98.17 | 97.64 | 97.83 | 98.00 | 97.15 | 97.85 | 86.86 | 83.47 | 83.05 | 95.74 | 94.92 | 95.21 | 98.32 | 97.75 | 98.11 | 95.42 | 94.19 | 94.41 |
| Accuracy | 99.04 | 98.77 | 98.87 | 99.38 | 99.12 | 99.34 | 97.18 | 96.21 | 96.32 | 98.69 | 98.46 | 98.53 | 98.92 | 98.56 | 98.78 | 96.61 | 95.57 | 95.92 |
| Cohen's Kappa | 97.52 | 96.81 | 97.06 | 97.52 | 96.64 | 97.46 | 85.29 | 81.33 | 81.00 | 94.97 | 94.01 | 94.34 | 97.52 | 96.70 | 97.21 | 95.58 | 94.24 | 94.69 |
| MCC | 97.50 | 96.80 | 91.09 | 97.52 | 96.63 | 97.39 | 85.28 | 81.30 | 81.05 | 94.99 | 94.03 | 94.33 | 97.52 | 96.81 | 97.08 | 95.60 | 94.24 | 94.70 |

Table 6. Evaluation Results for the Late-Stage Fusion (Majority Voting - 5 folds – 20 % test)

| Class | Fear in % | | | Joy in % | | | Relaxation in % | | | Sadness in % | | | Stress in % | | | All Classes in % | | |
|---|---|---|---|---|---|---|---|---|---|---|---|---|---|---|---|---|---|---|
| Classifier/ Metric | RF | DT | XGB | RF | DT | XGB | RF | DT | XGB | RF | DT | XGB | RF | DT | XGB | RF | DT | XGB |
| Precision | 73.10 | 89.96 | 89.82 | 87.36 | 89.95 | 89.09 | 87.82 | 89.86 | 63.06 | 97.96 | 89.98 | 92.13 | 80.83 | 89.98 | 95.83 | 85.41 | 89.95 | 85.99 |
| Recall | 90.59 | 89.98 | 97.43 | 82.42 | 89.97 | 95.23 | 42.83 | 89.81 | 42.89 | 63.83 | 89.96 | 92.46 | 95.43 | 89.98 | 96.84 | 75.02 | 89.94 | 84.97 |
| F1-score | 80.91 | 89.97 | 93.47 | 84.82 | 89.96 | 92.06 | 57.58 | 89.84 | 51.06 | 77.29 | 89.97 | 92.29 | 87.52 | 89.97 | 96.33 | 77.63 | 89.95 | 85.04 |
| Accuracy | 88.88 | 89.96 | 94.46 | 95.45 | 89.98 | 97.47 | 92.79 | 89.96 | 90.60 | 94.32 | 89.99 | 97.66 | 91.29 | 89.99 | 97.64 | 81.37 | 89.95 | 89.91 |
| Cohen's Kappa | 73.19 | 89.96 | 91.05 | 82.15 | 89.95 | 90.56 | 54.14 | 89.81 | 46.06 | 74.21 | 89.96 | 90.91 | 80.91 | 89.98 | 94.60 | 75.21 | 89.95 | 86.83 |
| MCC | 73.15 | 89.92 | 91.01 | 82.10 | 89.95 | 90.98 | 54.22 | 89.80 | 46.06 | 74.29 | 89.95 | 90.88 | 79.98 | 89.97 | 94.22 | 75.93 | 89.95 | 86.94 |

Table 4, which integrates body motion, physiological (EMG+GSR), and eye-tracking signals through feature selection, demonstrates that combining modalities substantially enhances emotion recognition performance. Overall results confirm that XGB and RF achieve balanced and reliable performance across emotions, with average F1-scores above 93%, strong precision (up to 93.5%), and high recall (93.0%). These values show that fused features capture complementary cues, allowing the classifiers to both identify emotional states accurately (precision) and cover a wide range of true cases (recall), particularly for high-arousal emotions like Fear and Sadness. Interestingly, DT reports extremely high scores across most metrics (e.g., All-Classes precision, recall, and F1 all around 96.9%, accuracy 97.8%). While these numbers suggest excellent recognition capability, such uniformly high values for a relatively simple classifier raise the possibility of overfitting to the fused feature space. This effect is particularly noticeable given DT's weaker performance in unimodal settings, which makes its sudden leap suspicious compared to the more stable gains of RF and XGB. In practice, this means that while DT appears to perform exceptionally, its results should be interpreted with caution, as it may be capturing noise or dataset-specific patterns rather than generalizable emotional markers.

    The second early-stage fusion in Table 5, based on autoencoder dimensionality reduction, shows that compressing multimodal features into a latent representation leads to highly stable and generalizable emotion recognition. Across the five classes (Fear, Joy, Relaxation, Sadness, Stress), both RF and XGB achieve strong overall scores, with RF slightly leading (All-Classes F1 = 95.4%, Accuracy = 96.6%) and XGB close behind (F1 = 94.4%, Accuracy = 95.9%). These results show that the autoencoder successfully preserves discriminative cues across modalities while reducing redundancy and noise. High precision (up to 97.9%) states that positive emotional states such as Joy and Sadness are rarely misclassified, while strong recall values (up to 98.7%) show that the models capture most true cases, even for subtler states like Relaxation. Importantly, unlike the feature-selection approach, where DT reported suspiciously inflated values, here the performance is more balanced across classifiers, proposing that the autoencoder representation supports generalization rather than overfitting.

    The late-stage fusion strategy based on majority voting provides stable but less powerful performance compared to early fusion methods. For the five emotional classes (Fear, Joy, Relaxation, Sadness, Stress), both RF and XGB show reasonably strong results, with XGB achieving an All-Classes F1 = 85.0% and Accuracy = 89.9%,

and RF producing high precision for certain emotions such as Stress (95.8%) but lower recall for subtler states like Relaxation (75.0%). This indicates that majority voting favors more distinct, high-arousal emotions while struggling to capture nuanced states, particularly Relaxation, where recall can drop below 50% in XGB. DT records unusually uniform values across all metrics (≈ 89.9%), which shows over-smoothing or bias from the voting mechanism, and therefore should be interpreted cautiously. Nevertheless, the overall Cohen's kappa and MCC values (~87–90%) confirm good agreement and consistent predictions across folds. In practical terms, majority voting provides robustness and classifier consensus, reducing the impact of individual model errors, but it does not fully exploit the richness of multimodal interactions. As a result, while effective for achieving balanced, interpretable outputs, late fusion remains less discriminative than early fusion approaches for fine-grained emotion recognition.

❖ **Data Analysis**

For data analysis, four categories are selected to save some space. Three modalities and the first early stage fusion by feature selection techniques. Figure 13 features the distribution across modalities of the proposed dataset. Also, Figure 14 represents the correlation analysis of the top ten features of our selected categories. Additionally, Figure 15 depicts feature importance analysis [87] of our selected portions.

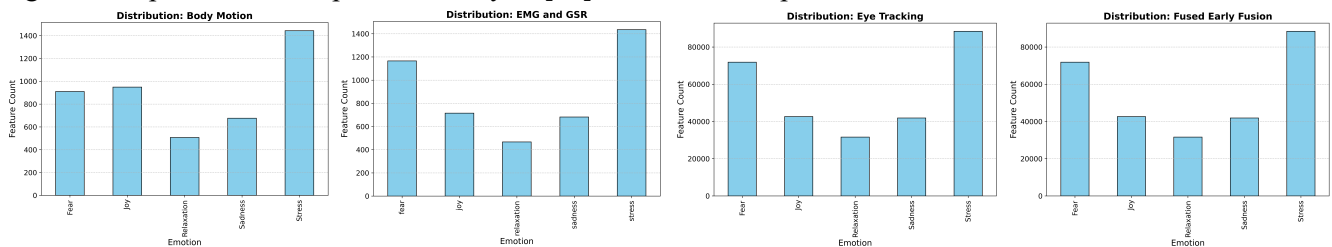

Figure 13. Feature distribution across modalities of our MVRS dataset

Figure 13 reveals notable differences in emotional representation. In the body motion and EMG+GSR datasets, Stress exhibits the highest feature counts, indicating pronounced behavioral and physiological variations during stressful states. Fear also shows consistently high representation, whereas Joy and Sadness appear moderately represented, and Relaxation demonstrates the lowest feature counts across these modalities, suggesting reduced variability in both motor and physiological signals during calm states. In contrast, the eye-tracking dataset produces substantially higher overall feature counts compared to the other modalities, with Stress and Fear again dominating, while Joy and Sadness form a mid-range cluster, and Relaxation remains the least represented. The fused early-stage features, which combine all modalities, follow a similar trend dominated by Stress and Fear, highlighting their stronger cross-modal signatures. Acquired patterns show potential class imbalance, with Stress and Fear contributing disproportionately more information than Relaxation, Joy, and Sadness, which may necessitate balancing strategies or advanced fusion techniques to make robust emotion classification performance.

The correlation analysis in Figure 14 shows modality-specific dependencies and redundancies. For body motion, strong negative correlations are observed between left-hand positional features (e.g., *HandTipLeft_PosX_rms* vs. *HandTipLeft_PosX_mean*, r ≈ –0.96), while several spectral features show high positive correlations (r > 0.9), showing overlapping information in spatial and frequency domains. In EMG and GSR, nearly all GSR-derived features (e.g., *GSR_Max*, *GSR_AUC*, *GSR_Min*, *GSR_Mean*) exhibit extremely high correlations (r > 0.95), stating their interdependence and showing potential redundancy if used simultaneously.

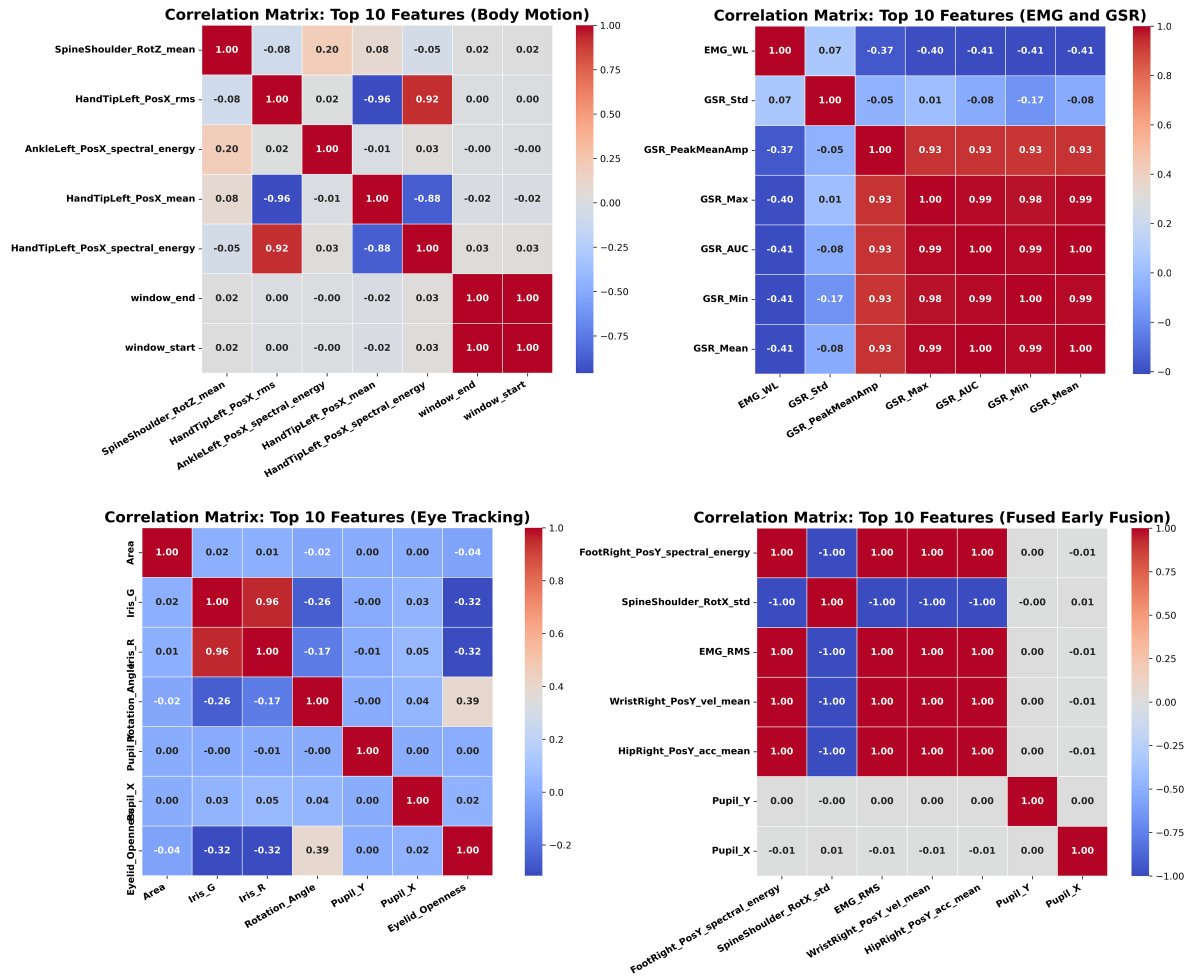

Figure 14. The correlation analysis of the top ten features of our selected categories

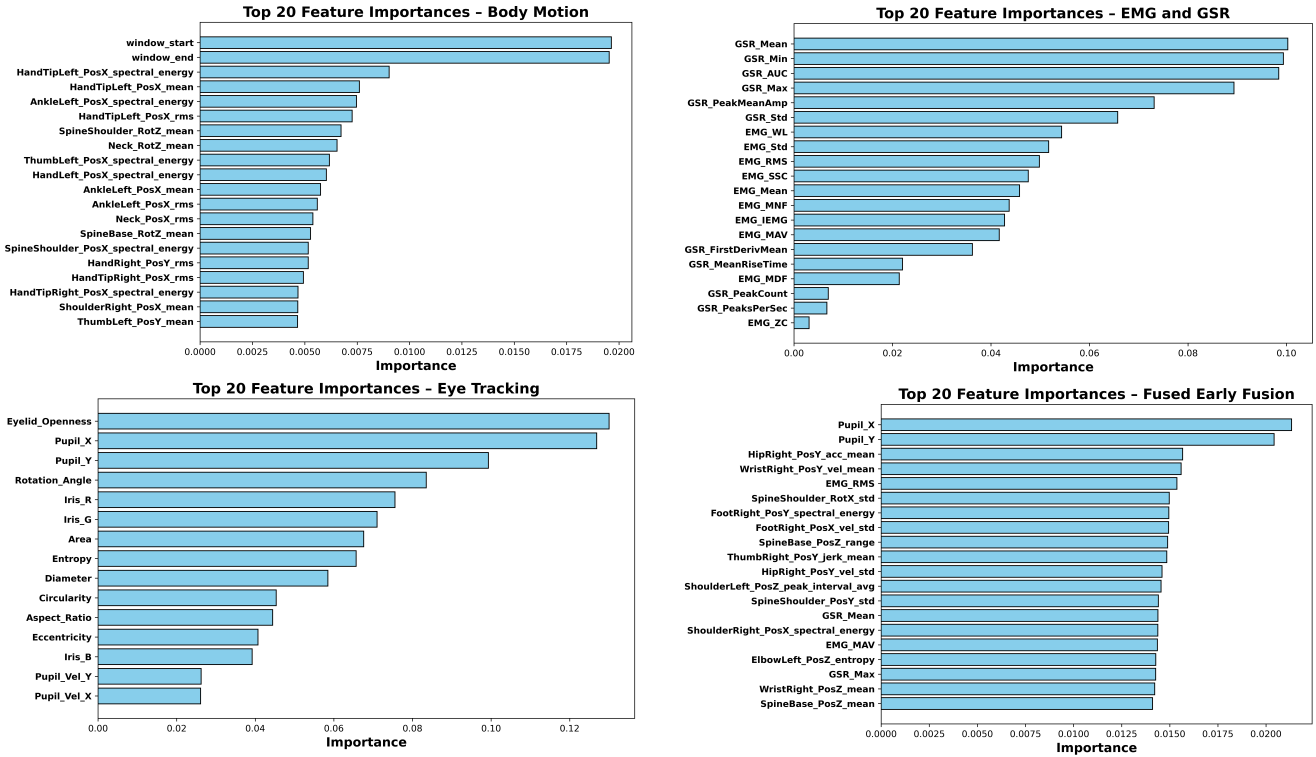

Figure 15. Feature importance analysis of our selected portions for our MVRS dataset

By contrast, eye-tracking features show more moderate correlations, with *Iris_C* and *Rotation_Angle* showing a strong positive relationship (r ≈ 0.96), while other features, such as *Pupil_X* and *Area,* remain largely uncorrelated, pointing to richer diversity and complementarity within this modality. In the fused early-stage features, clusters of variables drawn from different modalities (e.g., *FootRight_PosY_spectral_energy*, *SpineShoulder_RotX_std*, *EMG_RMS*) show nearly perfect correlations (r = ±1), suggesting that cross-modal fusion amplifies certain dominant patterns while also introducing redundancy. Achieved results emphasize the need for feature selection or dimensionality reduction in fusion scenarios to mitigate redundancy, while within unimodal analyses, careful handling of highly correlated physiological and kinematic features is essential to prevent overfitting and enable model generalization.

The feature importance plot in Figure 15 reveals that each modality contributes distinct yet complementary discriminative cues. For body motion, temporal markers (*window_start* and *window_end*) dominate, followed by kinematic features such as positional and spectral energy measures from the hands, ankles, and spine, indicating that temporal alignment and limb dynamics play key roles in differentiating emotional states. In EMG and GSR, skin conductance measures (*GSR_Mean*, *GSR_Min*, *GSR_AUC*, *GSR_Max*) are the most influential, supported by additional EMG-derived statistics, highlighting the strong discriminative capacity of physiological arousal signals. In eye tracking, the most important predictors are gaze and pupil-related measures (*Eyelid_Openness*, *Pupil_X*, *Pupil_Y*, *Rotation_Angle*), followed by iris color channels and shape descriptors, reflecting the sensitivity of ocular dynamics to emotional changes. Finally, in the fused early-stage features, the most influential variables include *Pupil_X* and *Pupil_Y* from eye tracking, alongside motion-based features (*HipRight_PosY_acc_mean*, *WristRight_PosY_vel_mean*, *SpineShoulder_RotX_std*) and physiological cues (*EMG_RMS*, *GSR_Mean*), demonstrating how multimodal fusion integrates complementary signals across behavior, physiology, and gaze to enhance emotion recognition. These results underscore the value of modality-specific strengths, kinematics for body motion, arousal points for EMG/GSR, and fine-grained gaze patterns for eye tracking, while confirming that fusion provides a more balanced and powerful feature set for classification. Figure 16 illustrates the cross-modality similarity analysis [88] of our modalities and the early-stage fusion data.

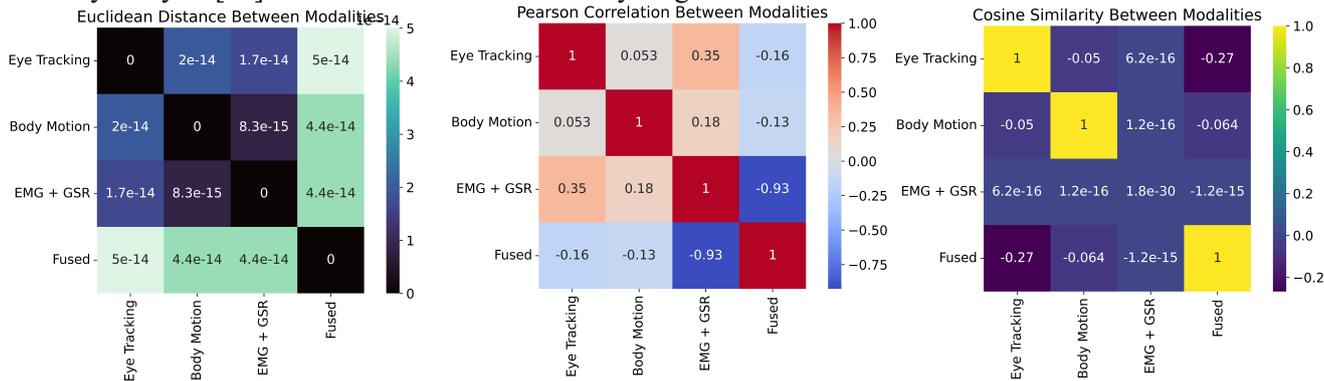

Figure 16. Cross-modality similarity analysis using Euclidean distance, Pearson correlation, and cosine similarity

The cross-modality similarity analysis provides insights into the degree of complementarity between feature spaces. The Euclidean distance matrix shows very small values across modalities, indicating that feature distributions are numerically comparable in scale, though not necessarily overlapping in structure. The Pearson correlation matrix reveals that eye tracking and EMG+GSR share the strongest positive relationship (r = 0.35), while body motion maintains only weak correlations with both eye tracking (r = 0.053) and EMG+GSR (r = 0.18), suggesting its feature space is more distinct. Interestingly, the fused features exhibit strong negative correlations with EMG+GSR (r = –0.93) and weaker negative associations with eye tracking (r = –0.16) and body motion (r = –0.13), representing how fusion introduces new combined patterns rather than simply aggregating modality-specific trends. The cosine similarity results further support these findings, showing that similarity between unimodal datasets remains close to zero, while the fused representation diverges from its individual components (e.g., negative similarity with eye tracking, –0.27). Figure 17 presents a PCA visualization of emotion feature distributions across modalities. Furthermore, Figure 18 depicts t-SNE visualization [89] of the same content. Figure 19 represents the UMAP [90] visualization of our data combinations.

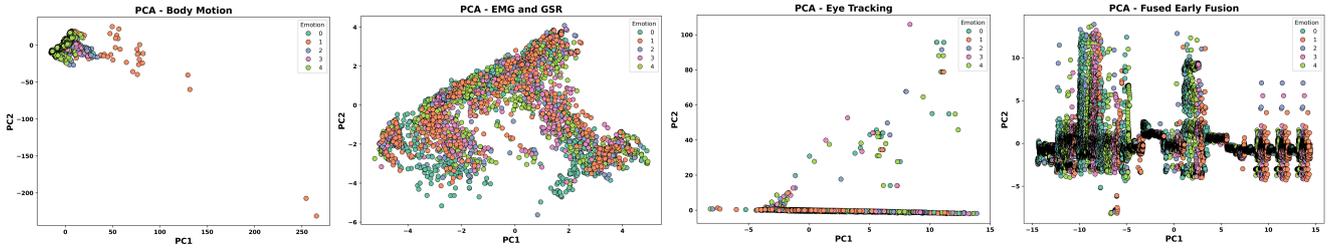

Figure 17. PCA visualization of feature distributions across modalities (0 → Fear, 1 → Joy, 2 → Relaxation, 3 → Sadness, 4 → Stress)

The PCA visualizations in Figure 17 provide insights into the separability of emotional states across modalities. For body motion, the data points form compact clusters with partial separation of Stress from other emotions, while Fear, Joy, and Sadness overlap considerably, and Relaxation remains embedded within the main cluster, indicating limited discriminability using motion alone. In EMG and GSR, the five emotions are widely intermingled, with only subtle structure visible in the projection, reflecting the continuous nature of physiological responses and the challenge of distinguishing emotions solely from arousal-related signals. The eye-tracking projection reveals a more distinct pattern, where Stress and Fear samples spread further along PC2, whereas Relaxation and Sadness cluster closer to the baseline, showing that gaze and pupil dynamics carry discriminative cues but remain partially overlapping. Finally, the fused early-stage features generate a more complex and structured distribution, with multiple dense groupings where Stress and Fear dominate several clusters, while Joy, Sadness, and Relaxation appear in more confined regions.

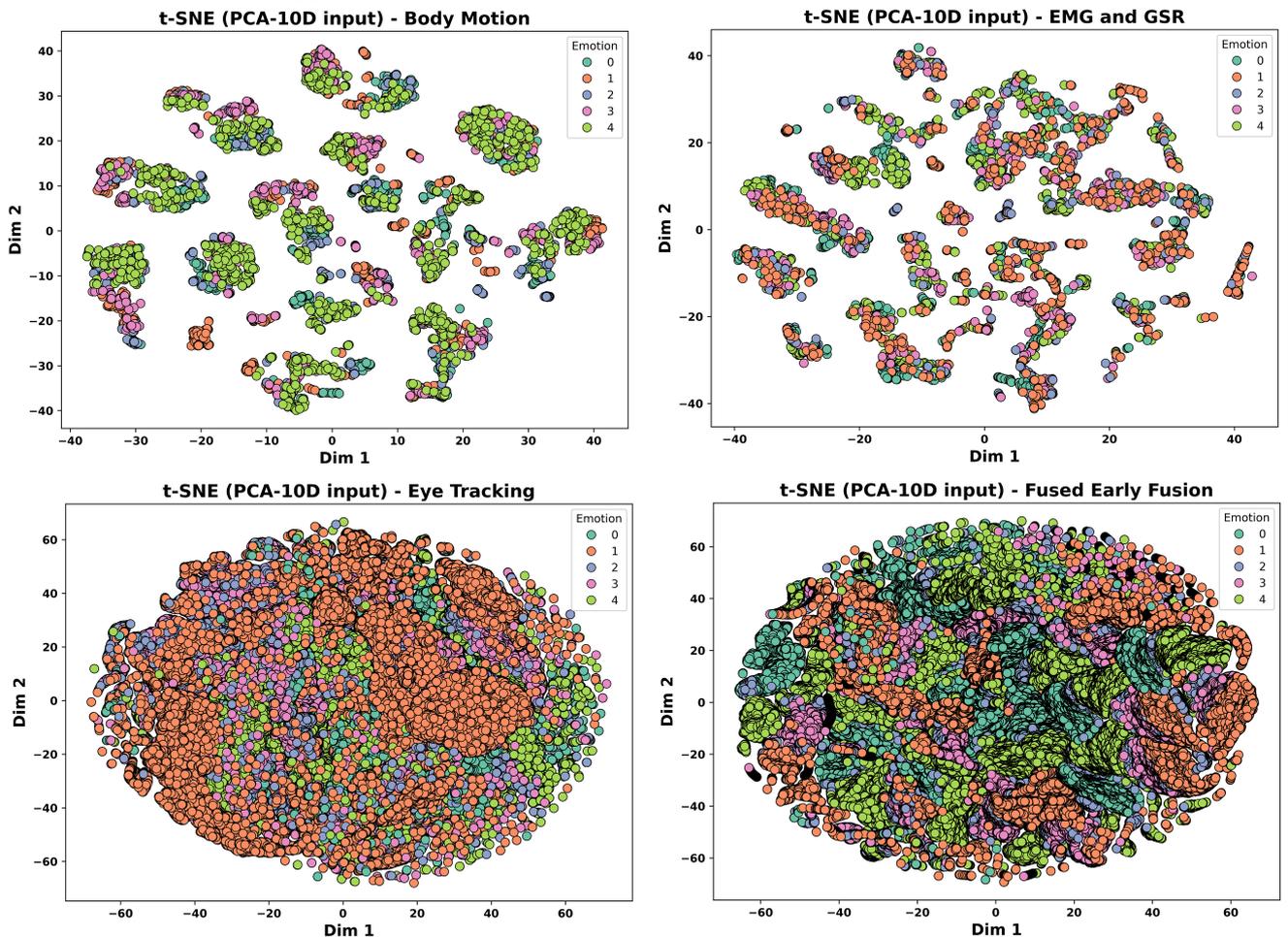

Figure 18. t-SNE visualization of body motion, EMG+GSR, eye tracking, and fused early fusion features of our MVRS dataset, showing partial clustering in unimodal spaces and improved separation in the fused representation. (0 → Fear, 1 → Joy, 2 → Relaxation, 3 → Sadness, 4 → Stress).

The t-SNE projections in Figure 18 provide a more nuanced view of the nonlinear separability of emotions across modalities. For body motion, the data forms multiple compact clusters with partial mixing of classes, but Stress and Fear samples appear more consistently separated into distinct subgroups compared to Joy, Sadness, and especially Relaxation, which remain dispersed within the clusters. In the EMG and GSR modality, the embedding reveals overlapping clouds with weak class separation, reflecting the continuous and highly correlated nature of physiological responses, where arousal-related signals blur the boundaries between emotional categories. The eye-tracking projection is dominated by a dense spread of Fear samples, with Joy, Sadness, and Relaxation sparsely interwoven and Stress occupying scattered regions, showing that gaze and pupil-based features capture subtle emotional variation but lack clear boundaries. In contrast, the fused early-stage features produce a richer manifold with structured patterns where Stress and Fear occupy more distinguishable regions, while Joy, Sadness, and Relaxation achieve improved separation compared to unimodal embeddings.

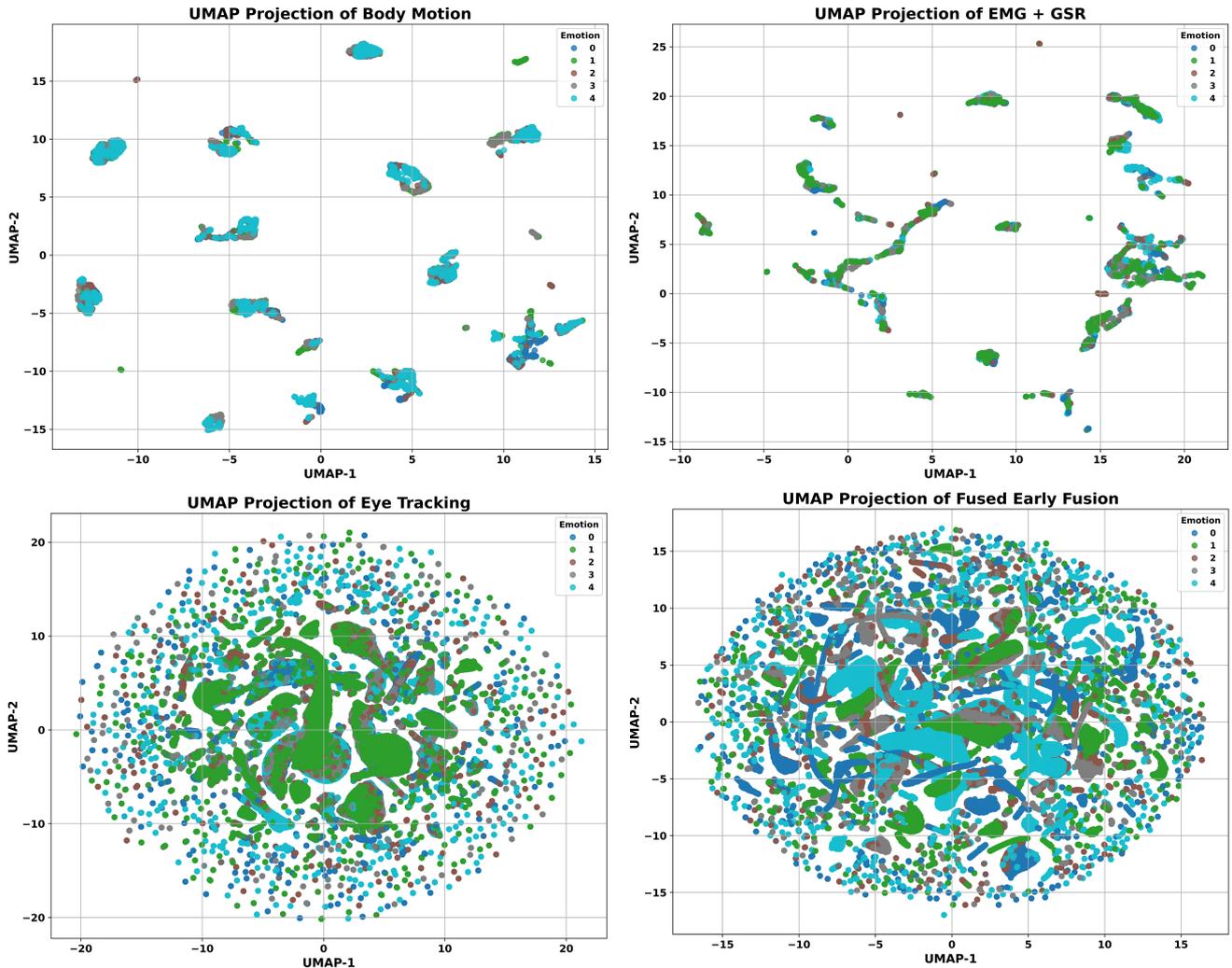

Figure 19. UMAP projections of unimodal and fused features, showing fragmented clusters in single modalities and clearer structure in fusion. (0 → Fear, 1 → Joy, 2 → Relaxation, 3 → Sadness, 4 → Stress).

The UMAP projections in Figure 19 reveal complementary insights into the emotional feature spaces. For body motion, the embedding forms multiple compact islands, with Stress and Fear showing partial grouping but Joy, Sadness, and especially Relaxation scattered across clusters, reflecting the heterogeneous nature of motion-based cues. In the EMG+GSR modality, the distribution is more elongated, with clusters of Stress and Fear emerging but still overlapping with Joy and Sadness, highlighting the continuous variability in physiological responses. The eye-tracking projection produces a dense circular manifold where Fear dominates the spread, while Relaxation and Sadness are intermixed in smaller regions, suggesting subtle differences in ocular dynamics that UMAP captures with less separation than motion-based features. The fused early-stage features create a richer and more evenly distributed structure, with multiple well-formed clusters where Stress and Fear appear more distinct, while Joy, Sadness, and Relaxation achieve improved visibility compared to unimodal projections. Figure 20 shows Log-SNR distributions [91] across modalities. Figure 21 illustrates Violin plots [92] of classification metrics across 30 folds for the second early-stage fusion (autoencoder + XGBoost).

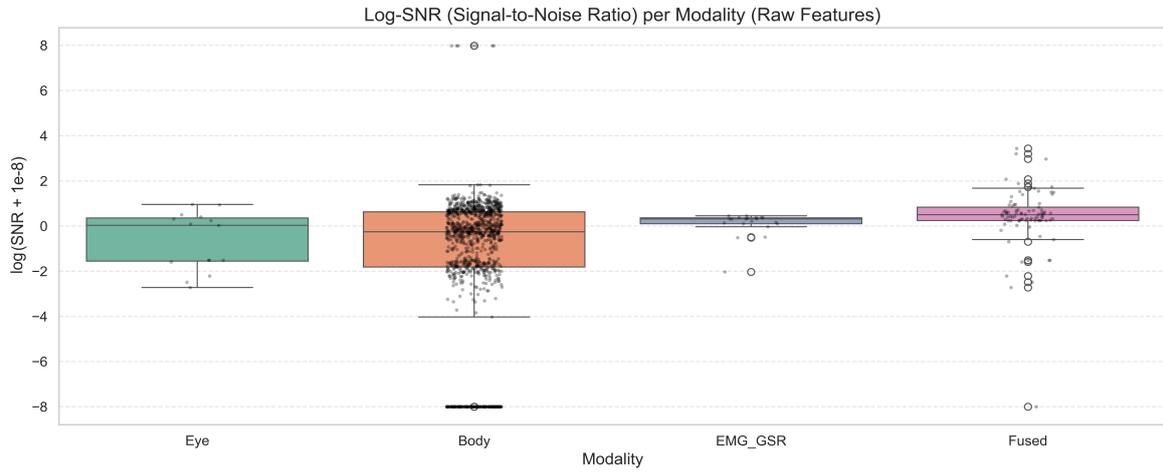

Figure 20. Log-SNR distributions across modalities, with moderate spread in eye tracking, wide variability in body motion, stable but low SNR in EMG+GSR, and a broader spread in the fused features.

The log-SNR distribution highlights substantial variability in signal quality across modalities. For the eye-tracking features, the SNR values cluster near zero with moderate spread, reflecting relatively balanced signal strength but with noticeable noise contributions. The body motion features display the widest spread, with several outliers reaching very low SNR values, indicating heterogeneous signal quality across joints and movements. The EMG+GSR features show the narrowest distribution, with most values tightly concentrated around zero to slightly positive SNR, suggesting more consistent but lower-amplitude signals compared to other modalities. In the fused dataset, the distribution is intermediate, broader than eye tracking and EMG+GSR but narrower than body motion, capturing contributions from all modalities without fully inheriting the extreme variability of motion signals. This pattern indicates that while fusion enhances overall representation richness, it also introduces noise from multiple modalities, emphasizing the need for feature selection or denoising to maximize discriminability.

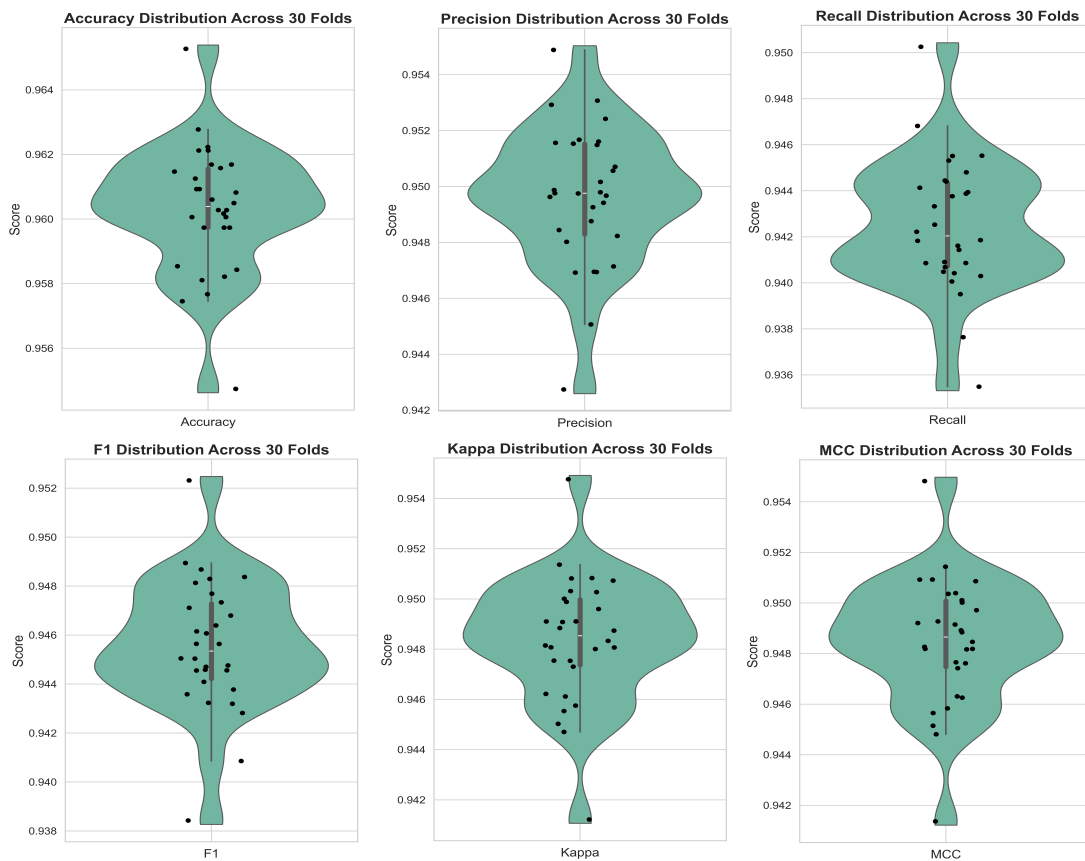

Figure 21. Violin plots of classification metrics across 30 folds for the second early-stage fusion (autoencoder + XGBoost), showing consistently high and stable performance across all measures (all classes).

The violin plots in Figure 21 summarize the classification performance of the second early-stage multimodal fusion (autoencoder-based) using the XGBoost classifier across 30 folds over all classes. The shape of each violin reflects the probability density of the metric values, where wider sections indicate more folds concentrated at that score range, and narrower sections indicate fewer occurrences. The black dots represent the individual fold results, making it possible to see the exact distribution of scores across all 30 evaluations. The vertical line within each violin marks the Inter Quartile Range (IQR), with the central white dot (or thick section depending on plotting style) representing the median, providing a robust summary of central tendency. The spacing between the dots conveys fold-to-fold variability: tightly packed dots around the median, as seen in accuracy and precision, indicate highly consistent performance, whereas slightly broader spacing in recall and F1 reveals modest variation across folds. The overall patterns show that accuracy is tightly clustered near 0.960, precision remains stable around 0.950, recall is slightly lower at ≈0.942–0.944, and F1 reflects the balance between the two at ≈0.946–0.948. Agreement-based measures, kappa and MCC, also remain densely centered near 0.949–0.951. Taken together, the violin shapes, dot distributions, and central lines show that the autoencoder-based fusion with XGBoost achieves consistently high and reliable results, with only minimal variability across folds.

❖ **Discussion (Answers to RQs)**

To address the RQs outlined in the introduction section, our findings provide several key insights. Regarding RQ1 (VR vs. 2D stimuli), the VR environment was effective in eliciting naturalistic emotional responses, with Stress and Fear consistently showing clearer separability across modalities and high classification scores (e.g., in body motion, XGBoost achieved 99.20% accuracy and 98.05% F1-score for Fear, compared to 97.86% accuracy for Relaxation, which was harder to distinguish). For RQ2 (role of synchronized multimodal data), the fused datasets outperformed unimodal baselines, with the first early-stage fusion achieving up to 99.45% accuracy (DT) and the second autoencoder-based fusion maintaining similarly high performance (≈99.38% accuracy, XGBoost), compared to unimodal accuracies which were lower and more variable (e.g., EMG+GSR maxed at 94.56% accuracy with RF, Eye Tracking at 97.51% with RF). Addressing RQ3 (early vs. late fusion strategies), early fusion clearly outperformed late fusion: while late fusion (majority voting) reached a maximum of 97.66% accuracy with XGBoost, early fusion exceeded 99%, showing that autoencoder-based integration produced more structured and discriminative feature spaces. For RQ4 (classifier comparison), XGBoost consistently delivered the best results across both unimodal and multimodal datasets, achieving the highest overall accuracy (99.38%) and F1-scores (≈98%), while Decision Trees were slightly less stable (dropping to 91–93% accuracy in unimodal EMG+GSR and Eye Tracking). Random Forest performed competitively but did not match XGBoost's peak reliability across folds. Finally, for RQ5 (most discriminative features per modality), feature importance analysis revealed that temporal and kinematic descriptors (e.g., window_start, positional energies) were most critical for body motion, skin conductance measures (e.g., GSR_Mean, GSR_Max) dominated in EMG+GSR, and pupil and gaze-based metrics (e.g., Pupil_X, Pupil_Y, Rotation_Angle) were most influential in eye tracking. Together, these results confirm that VR-elicited, synchronized multimodal signals, especially when fused early and analyzed with ensemble classifiers like XGBoost, enable highly accurate and reliable emotion recognition.

🔗 **Code Availability**

All codes for analyzing the dataset are available on GitHub as follows: Code

🔗 **Dataset Availability**

Our MVRS dataset is available on various platforms as follows: Kaggle, Zenodo, Figshare, and IEEE Data Port

🔗 **Conflict of Interest**

This research was independently conducted and fully funded by the authors under the Cyrus Intelligent Research initiative. The authors declare no conflicts of interest.

● **Conclusion**

The development and evaluation of the MVRS dataset have demonstrated its potential as a valuable benchmark for advancing multimodal emotion recognition in immersive environments. By employing VR stimuli, we successfully elicited naturalistic and ecologically valid emotional responses across five core affective states: Relaxation, Fear, Stress, Sadness, and Joy, providing richer dynamics than traditional 2D paradigms. Through synchronized acquisition of body motion, EMG+GSR, and eye-tracking signals, the dataset captures complementary behavioral, physiological, and ocular responses, offering a more holistic view of affective processes. Our extensive experiments showed that multimodal fusion significantly outperforms unimodal approaches, with early fusion, particularly when combined with autoencoder-based dimensionality reduction, yielding the highest classification performance (up to 99.38% accuracy and ~98% F1-score with XGBoost), compared to unimodal accuracies in the 91–97% range. The analyses further revealed that XGBoost consistently outperformed decision trees and random forest models, highlighting the importance of robust ensemble techniques in multimodal contexts. Feature importance studies underscored modality-specific discriminative cues, including temporal and kinematic descriptors in body motion, skin conductance metrics in EMG+GSR, and pupil/gaze dynamics in eye tracking. Complementary evaluations with PCA, t-SNE, UMAP, and log-SNR analyses confirmed that fusion produces more structured and balanced latent spaces while maintaining signal quality across modalities. Collectively, these results affirm that the MVRS dataset provides a reliable, comprehensive, and high-quality resource for the community, positioning it as a crucial step toward more naturalistic and accurate emotion recognition systems.

### ❖ Suggestions for Readers

Researchers and practitioners working with the MVRS dataset are encouraged to consider several aspects for maximizing its potential. First, while the dataset provides multiple synchronized modalities, careful preprocessing, and feature selection remain essential to handle redundancy and noise, particularly in fused scenarios where correlations can reach $r = \pm1$. Second, since VR-elicited responses often produce stronger variability in Stress and Fear compared to Relaxation and Sadness, readers should apply balancing strategies such as augmentation or class weighting to ensure fairness across categories. Third, violin plot analyses across 30 folds indicate high stability, but replication across different classifiers and architectures, such as deep learning frameworks, attention mechanisms, or graph neural networks, could produce additional insights. Finally, when deploying models trained on MVRS to real-world contexts, practitioners should be mindful of the ecological gap between controlled VR conditions and unconstrained environments, adapting their pipelines accordingly.

### ❖ Future Work

Moving forward, we plan to extend the MVRS dataset in several directions. On the data collection side, we aim to increase the sample size and participant diversity to improve generalizability, as well as incorporate additional physiological modalities (such as EEG, respiration, or heart-rate variability) to enrich affective coverage. Stimuli design will also be enhanced by exploring adaptive VR scenarios that tailor emotional intensity dynamically to participant reactions. Methodologically, we intend to explore advanced fusion strategies such as attention-based fusion, deep multimodal transformers, and contrastive representation learning, as well as domain adaptation techniques to bridge the gap between VR and real-world settings. Another direction involves evaluating the dataset using eXplainable AI (XAI) approaches, providing a clearer interpretation of modality contributions for both researchers and clinicians. Lastly, we plan to make the dataset more accessible by offering standardized baselines, pretrained models, and benchmarking protocols, verifying that the MVRS resource continues to serve as a foundation for innovation in multimodal emotion recognition research.

## ● Annex

To complement the main evaluation results, the following annexes present representative temporal trends of selected features from each modality in our proposed MVRS dataset. These plots illustrate how raw feature values evolve over time or across 1000 sampled points, emphasizing the variability, stability, and discriminative patterns captured by different sensors. Annex A focuses on body motion features, showcasing positional, rotational, spectral, and entropy-based measures that reflect both subtle postural adjustments and sudden movement bursts, which are often tied to emotionally charged expressions such as stress or fear. Annex B presents EMG and GSR features, demonstrating the contrast between fast, high-frequency muscle activations and slower, cumulative electrodermal responses; together, these features reveal how the body simultaneously encodes short-lived muscular reactions and longer-term arousal states in response to VR stimuli. Annex C highlights eye-tracking features, including pupil size, iris metrics, eyelid openness, and gaze dynamics, which capture both gradual ocular changes, such as pupil dilation under stress, and rapid saccadic events associated with heightened attentional shifts. Taken as a whole, the annexes provide a more tangible, fine-grained view of how multimodal features behave across time, reinforcing their complementary role in multimodal fusion and confirming the dataset's ability to capture the complex dynamics of emotion expression. Beyond serving as supplementary evidence, these visualizations also offer practical insights for future researchers, as they illustrate which signals remain stable, which fluctuate sharply, and how such differences can guide the design of preprocessing pipelines, feature selection strategies, and classifier architectures in multimodal emotion recognition.

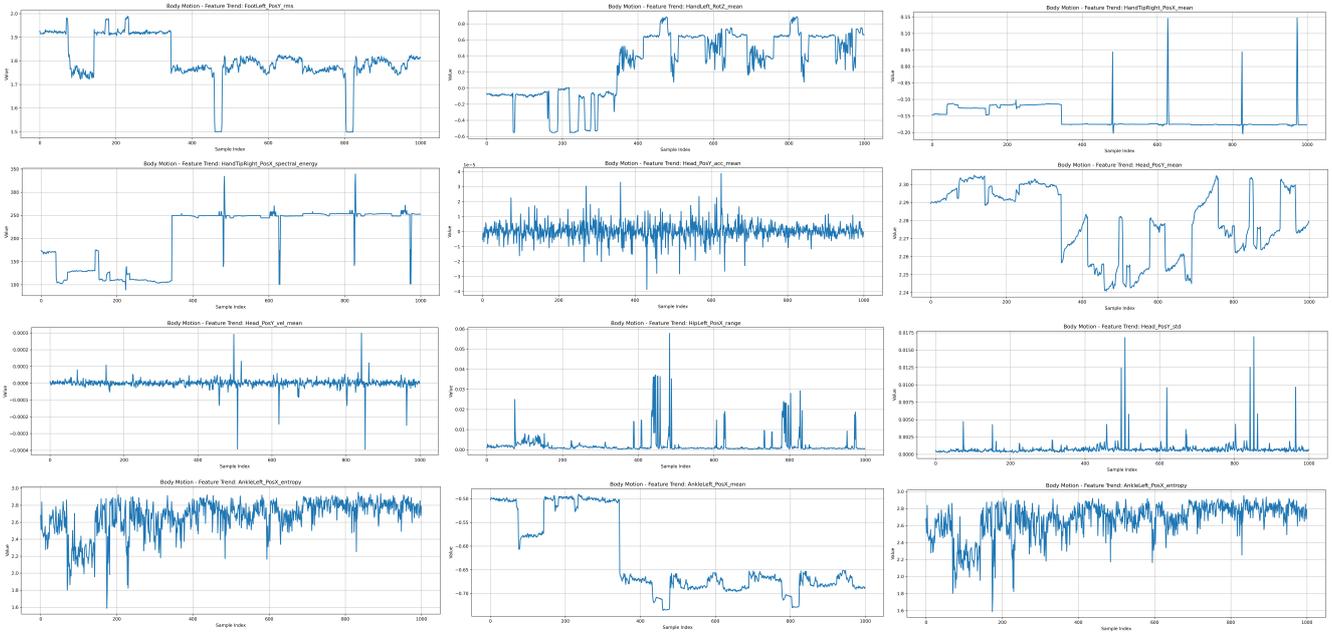

**Annex A.** *Representative temporal trends of selected body motion features (e.g., positional, rotational, spectral, and entropy measures) across 1000 samples, illustrating variability from smooth postural shifts to sharp movement bursts.*

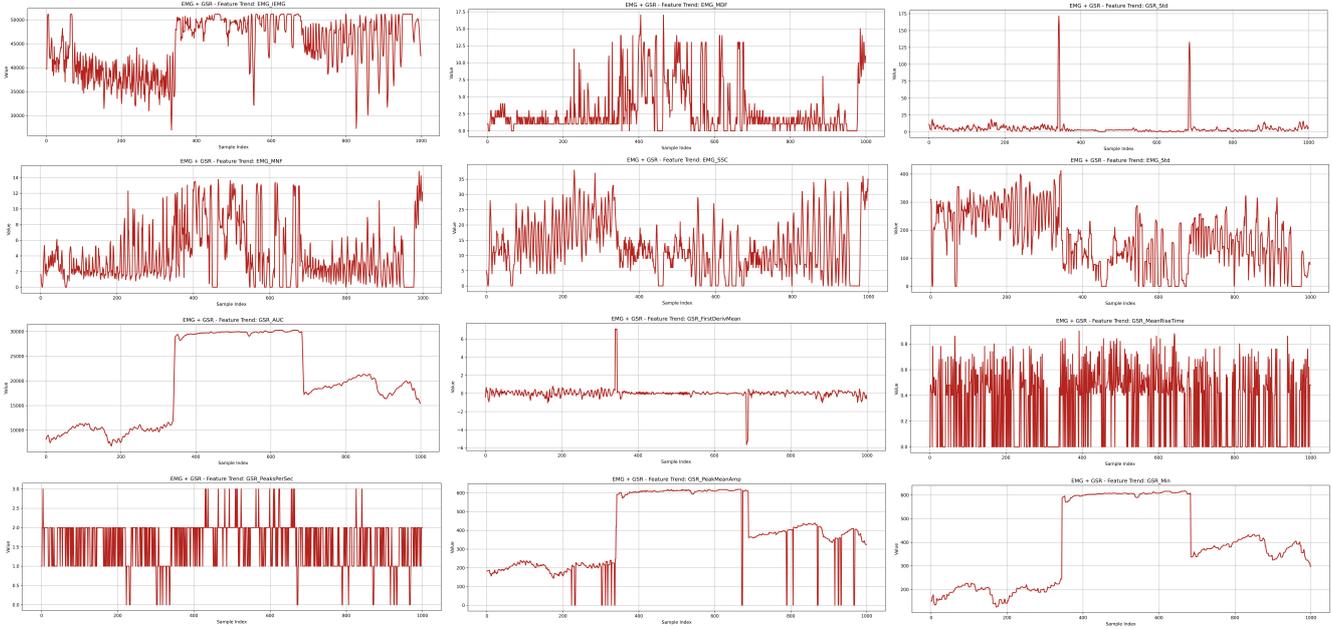

**Annex B.** *Representative temporal trends of selected EMG and GSR features (e.g., frequency-domain, statistical, and arousal indicators) across 1000 samples, showing the interplay between fast muscle activations and slower electrodermal responses.*

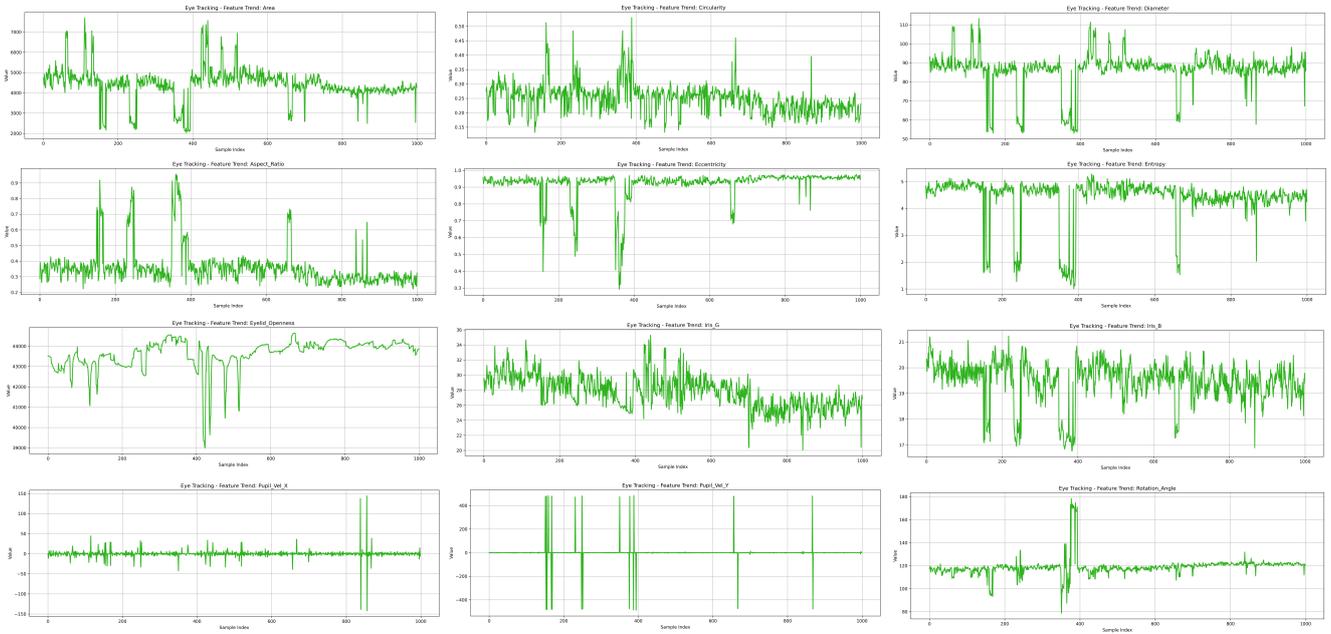

**Annex C.** *Representative temporal trends of selected eye-tracking features (e.g., pupil size, eyelid openness, iris metrics, and gaze dynamics) across 1000 samples, capturing both gradual changes in ocular state and rapid saccadic events.*

The selected body motion features in Annex A show the diversity of kinematic and positional dynamics over time. Features such as FootLeft_PosY_rms and HandTipRight_PosX_spectral_energy reveal fluctuations and periodic changes, capturing energy variations associated with movement intensity. Positional descriptors like HandLeft_RotZ_mean, Head_PosY_mean, and AnkleLeft_PosX_mean demonstrate gradual shifts interspersed with sharp transitions, reflecting posture adjustments and limb movements characteristic of emotion-related body expression. Velocity and acceleration measures (e.g., Head_PosY_vel_mean, Head_PosY_acc_mean) remain relatively stable but show sporadic spikes, indicating subtle motion bursts or micro-adjustments in head position. Entropy-based features, including AnkleLeft_PosX_entropy, exhibit higher variability, reflecting the irregularity and unpredictability of certain joint trajectories under different emotional states. Overall, these trends depict how body motion provides a rich set of temporal patterns, ranging from smooth positional drifts to sharp bursts and irregular fluctuations, which together form discriminative cues for emotion recognition in VR environments.

The EMG and GSR features in Annex B present the interplay between muscle activity and skin conductance responses as indicators of arousal and affective states. EMG-derived measures such as IEMG, MNF, MDF, and SSC display high variability and frequent oscillations, stating dynamic changes in muscular activation intensity, frequency content, and signal complexity over time. Features like EMG_Std and EMG_SSC capture burst-like behaviors, where peaks coincide with periods of heightened muscular tension, consistent with stress- or fear-related responses. In contrast, GSR-related measures such as AUC, PeakMeanAmp, and Min exhibit longer-range trends, with step-like increases followed by gradual decreases, capturing the slower dynamics of electrodermal activity as the sympathetic nervous system responds to stimuli. Rapid fluctuations in features such as PeaksPerSec and MeanRiseTime highlight moment-to-moment variability in sweat gland activation, providing fine-grained resolution of arousal events. Collectively, the EMG features capture fast, high-frequency muscle changes, while GSR features provide slower, cumulative indicators of physiological arousal, and their combination offers a powerful window into emotion recognition, with complementary temporal scales spanning milliseconds to several seconds.

The selected eye-tracking features in Annex C capture both geometric properties of the pupil/iris and dynamic changes in gaze behavior. Shape-related features such as Area, Diameter, Aspect_Ratio, Circularity, and Eccentricity display fluctuations over time, showing natural pupil dilation/constriction and morphological variability during emotional engagement. Entropy and Euclidean Openness highlight irregularities in eyelid movements, where sudden drops indicate blinking or partial closures, consistent with stress- or fatigue-induced modulation of gaze. Iris-based color channel intensities (e.g., Iris_G, Iris_B) demonstrate finer-scale variability that indirectly captures lighting changes or micro-adjustments in gaze direction. Dynamic indicators such as Pupil_Vel_X, Pupil_Vel_Y, and Rotation_Angle show sharp bursts amidst otherwise stable baselines, signaling rapid saccadic shifts or head–eye coordination events, often associated with heightened cognitive or emotional processing. In summary, these temporal traces illustrate that eye-tracking features provide both slow-varying cues (pupil size, iris measures) and fast transients (saccades, blinks), making them critical for capturing subtle affective states within VR environments.

- *References*